\documentclass[preprint,12pt]{elsarticle}

\usepackage{url}
\usepackage{multirow}
\usepackage{subfigure}
\usepackage{booktabs}
\UseRawInputEncoding

\usepackage{amssymb}
\usepackage{amsmath}

\journal{}

\begin{document}

\begin{frontmatter}



\title{Supervised Visual Docking Network for Unmanned Surface Vehicles Using Auto-labeling in Real-world Water Environments}

\author[label1]{Yijie Chu}
\author[label2]{Ziniu Wu} 
\author[label3]{Yong Yue} 
\author[label3]{Eng Gee Lim} 
\author[label4]{Paolo Paoletti}
\author[label3]{Xiaohui Zhu}
\ead{xiaohui.zhu@xjtlu.edu.cn}
\affiliation[label1]{organization={School of Information Science and Engineering, Hebei University of Science and Technology},
            addressline={26 Yuxiang Street, Yuhua District}, 
            city={Shijiazhuang, Heibei Province},
            postcode={050018}, 
            country={P.R. China}}
            
\affiliation[label2]{organization={School of Civil, Aerospace and Design Engineering, University of Bristol},
            addressline={Queens Road}, 
            city={Bristol},
            postcode={BS8~1QU}, 
            country={UK}}
            
\affiliation[label3]{organization={School of Advanced Technology, Xi’an Jiaotong-Liverpool University},
            addressline={111 Ren’ai Road, Suzhou Industrial Park}, 
            city={Suzhou, Jiangsu Province},
            postcode={215123}, 
            country={P.R. China}}
            
\affiliation[label4]{organization={Department of Mechanical, Materials and Aerospace Engineering, University of Liverpool},
            addressline={Foundation Building,
Brownlow Hill}, 
            city={Liverpool},
            postcode={L69 3GH}, 
            country={UK}}

\begin{abstract}
Unmanned Surface Vehicles (USVs) are increasingly applied to water operations such as environmental monitoring and river-map modeling. It faces a significant challenge in achieving precise autonomous docking at ports or stations, still relying on remote human control or external positioning systems for accuracy and safety which limits the full potential of human-out-of-loop deployment for USVs.
This paper introduces a novel supervised learning pipeline with the auto-labeling technique for USVs autonomous visual docking. Firstly, we designed an auto-labeling data collection pipeline that appends relative pose and image pair to the dataset. This step does not require conventional manual labeling for supervised learning. Secondly, the Neural Dock Pose Estimator (NDPE) is proposed to achieve relative dock pose prediction without the need for hand-crafted feature engineering, camera calibration, and peripheral markers. Moreover, The NDPE can accurately predict the relative dock pose in real-world water environments, facilitating the implementation of Position-Based Visual Servo (PBVS) and low-level motion controllers for efficient and autonomous docking.
Experiments show that the NDPE is robust to the disturbance of the distance and the USV velocity. The effectiveness of our proposed solution is tested and validated in real-world water environments, reflecting its capability to handle real-world autonomous docking tasks. 
The dataset and experimental videos are available at our project homepage: \url{https://sites.google.com/view/usv-docking/home}.
\end{abstract}

\begin{graphicalabstract}
\includegraphics[width=1\linewidth]{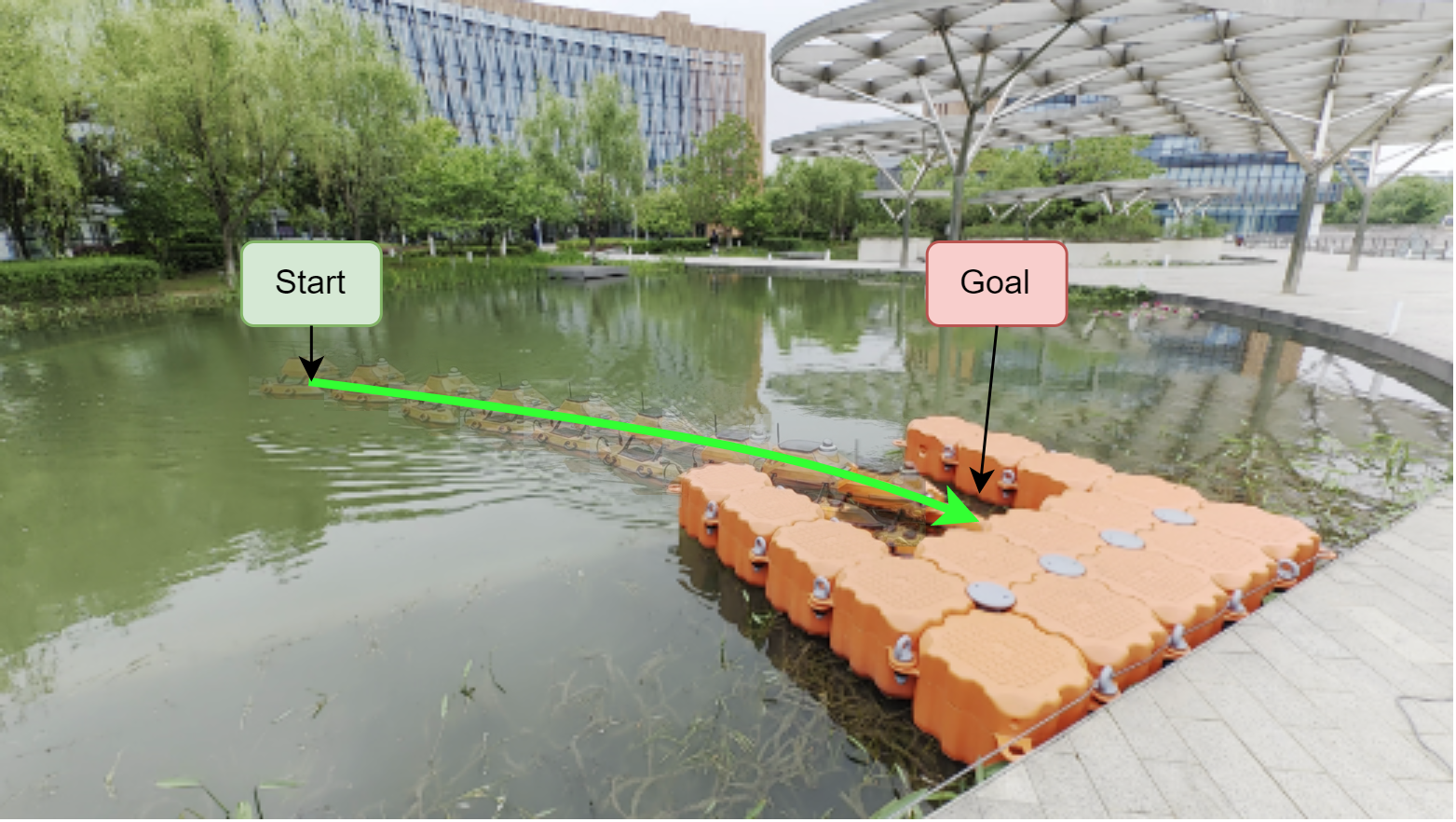}
\includegraphics[width=1\linewidth]{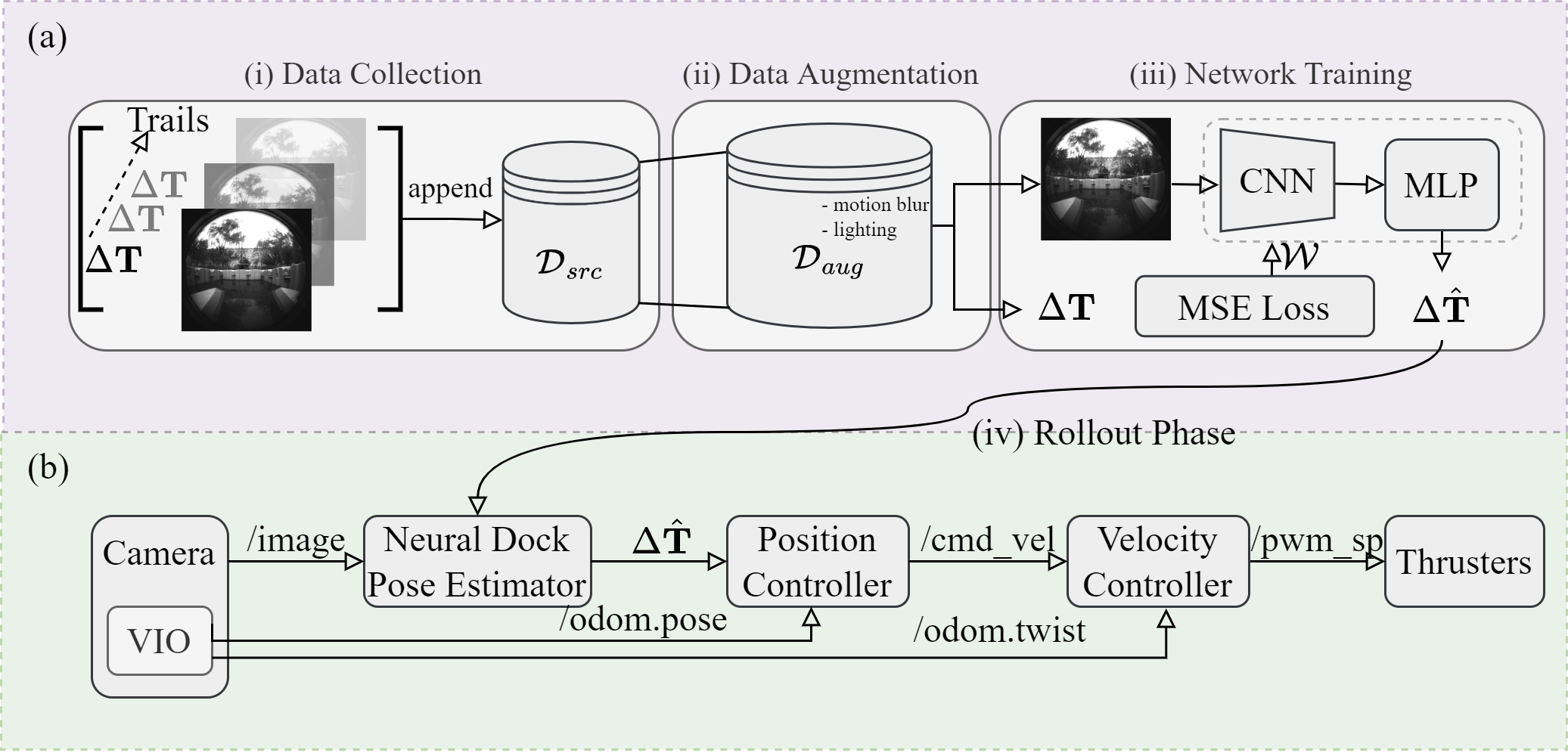}
\end{graphicalabstract}

\begin{highlights}
\item We designed an efficient auto-labeling pipeline eliminating the need for human labeling for data collection for supervised learning.
\item We developed a Neural Dock Pose Estimator (NDPE) proposed to achieve relative dock pose prediction without the need for hand-crafted feature engineering, camera calibration, and peripheral markers.
\item We implemented the Position-Based Visual Servo (PBVS) and low-level controller for the USV motion control.
\item We conducted several field experiments in real-world water environments to validate the performance of the proposed visual docking framework.
\end{highlights}


\end{frontmatter}


\section{Introduction}

\begin{figure}[htp]
	\centering        
        \includegraphics[width = 1\linewidth]{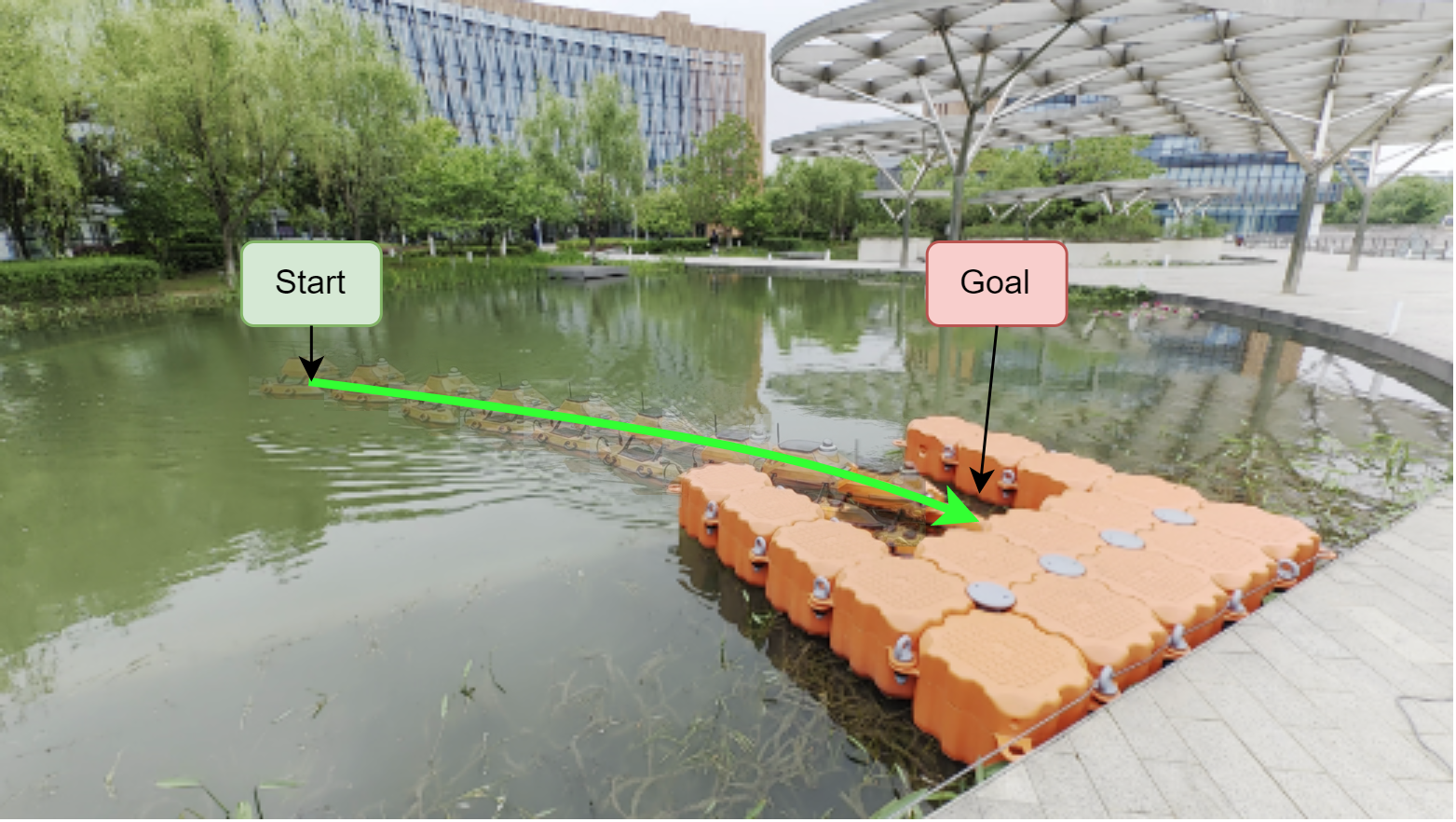}
	\caption{Illustration of a real-world autonomous docking task. The USV starts at the pre-docking area and then autonomously navigates to the dock.}
	\label{fig:concepts}
\end{figure}

Unmanned Surface Vehicles (USVs) have become increasingly integrated into applications in surveillance, environmental monitoring, and cargo transportation \cite{liu2016unmanned,glaviano2022management,ubina2022review}. Their conventional operation often necessitates human control, particularly for re-docking after autonomous missions. Autonomous docking aims to significantly enhance the functionality and application range of USVs, particularly for fully unmanned deployments \cite{aune2019development,hansen2006autonomous}. Figure \ref{fig:concepts} illustrates a scenario of the real-world autonomous docking task. The USV begins at the start which is near the dock and then autonomously navigates to the dock. To achieve precise autonomous docking, visual servoing (VS) techniques are employed.

\subsection{Visual Servo for Autonomous Docking}

In the autonomous docking task for USVs, Visual Servo (VS) techniques play a crucial role, utilizing visual markers or other features captured in images to guide the USV relative to the dock \cite{kim2012toward}. Position-Based Visual Servo (PBVS) and Image-Based Visual Servo (IBVS) are two primary approaches in visual servoing, each with distinct advantages and limitations \cite{corke2011robotics,lang2016application,peng2020comparing}. PBVS, which relies on accurate 3D models and pose estimation, offers significant benefits for tasks requiring precise 3D positioning and control. Its robustness to image noise, ability to decouple position and orientation control, and global convergence properties make it ideal for complex applications, despite its higher computational demands and the need for meticulous calibration. In contrast, IBVS directly utilizes 2D image features, simplifying implementation and reducing computational load, but it is more sensitive to noise and calibration errors and prone to local minima. Given our objectives of precision and reliability in dynamic environments, PBVS is the more suitable choice, providing the accuracy and control essential for our applications.

\begin{figure}[htp]
    \centering
    \includegraphics[width=0.7\linewidth]{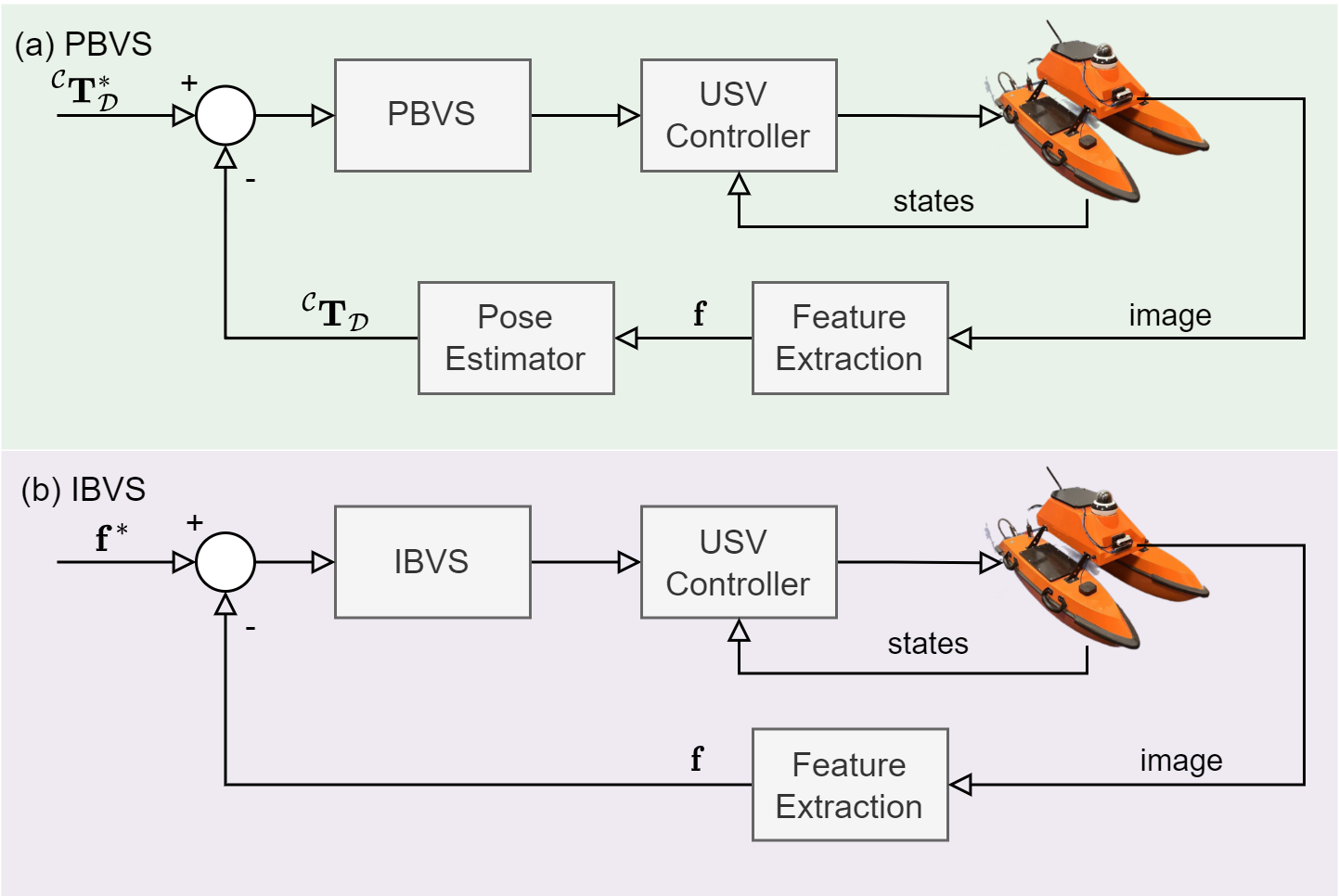}
    \caption{Block diagrams of visual-servo methods.}
    \label{fig:vs}
\end{figure}

\subsubsection{Position-Based Visual Servo (PBVS)}
In Figure \ref{fig:vs}, PBVS employs visual features observed through a calibrated camera and a known geometric model of the target to estimate its pose relative to the camera. The USV then maneuvers towards this pose, with control operations executed in the task space, commonly referred to as $\mathbf{SE(3)}$. Although effective, PBVS is computationally intensive and heavily depends on precise camera calibration and the accuracy of the geometric model of the target. For example, Dunbabin et al. \cite{dunbabin_vision-based_2008} detailed the creation of a monocular vision-based autonomous surface vehicle designed for coordinated docking maneuvers. The system architecture integrates two processor units: one for vehicle control using a virtual force-based docking strategy, and another for vision-based target segmentation and tracking. 

On the other hand, Volden et al. \cite{volden_vision-based_2022} showcased a stereo camera-based PBVS for USV docking. This system uses a YOLOv3 object detection model that takes stereo images as input. However, this method requires extensive manual labeling using bounding boxes, feature engineering, and stereo-view geometry for pose estimation.

Cai et al. \cite{cai_long-range_2023} introduced an position-based path planning algorithm for USVs, utilizing long-range Ultra-Wideband (UWB) positioning to automate docking. This approach combined a median filter (MF) with an extended Kalman filter (EKF) to refine the USV's positional accuracy in UWB-based coordinates. Their findings demonstrated that UWB-based positioning offers superior accuracy compared to traditional GPS, making it ideal for GPS-denied environments. In sea trials, this high-precision positioning allowed for the USV to navigate back to the dock following a meticulously designed iterative Dubins curve.

Meanwhile, Wang and He \cite{wang_extreme_2021} developed a homography-based visual-servo system that tackles the dynamic-level visual servo challenges. By incorporating a model-free neural network to estimate nonlinear dynamic terms, this system compensated for homography-based error dynamics. The limitation of this work is that the proposed neural network is implicitly specified to certain USV configurations.

\subsubsection{Image-Based Visual Servo (IBVS)}

IBVS, by contrast, operates directly with image features, eliminating the need for pose estimation and performing control in the image coordinate space $\mathcal{R}^2$. This method defined the desired camera pose implicitly by the feature values at the goal pose, presenting a complex control challenge due to the highly nonlinear relationship between image features and camera pose \cite{shademan2010robust}. 

Jiang et al. \cite{jiang2024image} introduced an innovative IBVS control strategy for USVs to manage the challenges posed by wave perturbations.  The primary contributions of their work include the development of a perturbations-observer-based controller that compensates for wave-induced roll and pitch, allowing for accurate positioning based solely on image coordinates without requiring specific pose information of the USV.  Additionally, they provided a stability proof demonstrating the global asymptotic stability of the proposed IBVS system, ensuring reliable operation under various environmental disturbances.
In vehicle docking task, Yahya et al. \cite{yahya2017image} developed an IBVS method for autonomous underwater vehicles (AUVs) to perform docking maneuvers using visual information from five distinguishable target features.  This method improves docking accuracy by relying solely on the motion of image features, eliminating the need for precise pose estimation, and demonstrating robustness in complex underwater conditions.

The limitations of IBVS include its direct dependence on image features for control, making it particularly sensitive to image noise, occlusions, and lighting variations. Additionally, IBVS can face controlling stability issues, especially when features experience significant movement or when the image Jacobian (which relates image feature velocities to camera velocities) becomes poorly conditioned, potentially leading to instability in the control system \cite{allibert2010predictive}.
\subsection{Neural-based Control for USVs}

In the field of maritime navigation, research in motion control and path planning for USVs has become increasingly significant. Researchers have developed a range of motion control strategies \cite{liu2019adaptive,er2023intelligent}, from basic to advanced, to enhance the precision and adaptability of autonomous maritime operations \cite{chu2022pk}. Particularly, the integration of neural network models has changed the control, enabling the learning of behavioral patterns from traditional controllers and utilizing the networks' robust generalization capabilities to surpass other methods \cite{dong2019heading,yu2019finite}.

For instance, Liu et al. \cite{liu2018bounded} crafted a neural network controller that incorporates an extended state observer (ESO) to refine control mechanisms. To further improve performance and avoid input saturation, Liu et al. \cite{liu2021robust} introduced an adaptive self-structuring neural network. This innovative approach approximates model uncertainties and external disturbances effectively. Additionally, by dynamically adjusting the number of neurons, this method allows for real-time optimization of the neural network's structure, which not only enhances performance but also reduces the computational load. Park et al. \cite{park2017neural} developed a neural network-based observer designed to accurately estimate velocity in uncertain conditions. Leveraging the state estimates provided by this observer, they then designed an output feedback controller capable of handling both input saturation and underactuated problems, further advancing the capabilities of neural network applications in control systems. Elhaki and Shojaei \cite{elhaki2021robust} introduced a novel saturated dynamic surface controller by integrating adaptive neural networks with robust adaptive controllers. This approach employed neural networks to counteract the non-linearities caused by actuator saturation, effectively reducing the risk of saturation in USV actuators.

Neural networks significantly enhance the motion control and path planning of USVs by leveraging their robust generalization capabilities and the ability to learn from traditional controllers. Neural network-based approaches, such as adaptive self-structuring neural networks and neural network-based observers, have been developed to address challenges like input saturation, model uncertainties, and external disturbances. By dynamically adjusting the network structure in real-time and integrating traditional control mechanisms, these methods achieve superior control effects, improving both performance and computational efficiency. These innovations demonstrate the robust capabilities of neural networks in advancing USV control systems and enhancing the precision and adaptability of autonomous maritime operations.

\subsection{Research Challenges}
The primary challenge in autonomous USV docking lies in navigating dynamic and unpredictable water environments, which present complexities such as changing weather conditions, water dynamics, and obstacles \cite{wu2024overview}. These challenges require sophisticated navigation, guidance and control (NGC) algorithms \cite{mousazadeh2018developing}. GPS provides global positioning information for the navigation of USVs. In particular, GPS-based methods are reliable most of the time, however, occasionally glitching or lagging is likely to be encountered, which may lead to inaccurate state estimation. Precise navigation tasks such as autonomous docking, require a high accuracy of local navigation methods. This can be addressed through visual servo (VS) for a closed-loop feedback control \cite{dunbabin_vision-based_2008,volden_vision-based_2022,cai_long-range_2023,wang_extreme_2021}. PBVS determines the pose of the target object relative to the camera using observed visual features and a calibrated camera\cite{corke2011robotics}. However, PBVS often requires peripheral labels, such as different colors and shape tags \cite{bingham19toward}, or ArUco markers \cite{aruco2014}. The essential difference between IBVS and PBVS is that the relative pose of the target is not estimated. The relative pose is implicit in the image feature values. The control problem can be expressed in terms of image coordinates so that the changing feature points in the image implicitly change the pose of the camera. IBVS has changed the problem from pose estimation to control of points in the image. If conventional VS is used, the worst-case scenario requires only one observation of the target to estimate the target pose, and then the structured controller performs a long horizon of control to reach the goal. Therefore, traditional VS approaches, including both PBVS and IBVS,  often rely heavily on manual feature engineering \cite{dunbabin_vision-based_2008}, extensive data labeling \cite{volden_vision-based_2022}, and high-precision localization technology \cite{cai_long-range_2023}, all of which still demand significant human effort.

To address these challenges, we explore the use of novel supervised learning-based techniques to enable visual docking in USVs with minimal human intervention. The research aims to assess how the proposed approach can support accurate and dependable autonomous docking in the dynamic conditions of real-world water environments. Through the auto-labeling design of data collection, autonomous labeling with minimal human involvement, and the use of a simple but effective neural dock pose estimator (NDPE), these methods can improve operational efficiency without compromising accuracy. Furthermore, research also contributes to the broader goal of advancing robot learning technology for practical real-world applications and operations for USVs. 

\subsection{Contributions}
The novelties and contributions of this work are:
\begin{itemize}
    \item We designed an efficient auto-labeling pipeline eliminating the need for human labeling for data collection for supervised learning.
    \item We developed a Neural Dock Pose Estimator (NDPE) proposed to achieve relative dock pose prediction without the need for hand-crafted feature engineering, camera calibration, and peripheral markers.
    \item We implemented the Position-Based Visual Servo (PBVS) and low-level controller for the USV motion control.
    \item We conducted several field experiments in real-world water environments to validate the performance of the proposed visual docking framework.
\end{itemize}

Section \ref{section2} concentrates on developing a novel supervised learning-based positional visual docking pipeline for autonomous visual docking. This entails conceptualizing and implementing a model capable of adapting to various real-world docking scenarios. Subsequently, the paper assesses the performance of our pipeline within a real-world water environment, where a range of docking situations are tested in Section \ref{sec:experimentsandresults}. Finally, we discuss the limitations of our approach and further research direction in Section \ref{section conclusion}.
\section{Supervised Learning-based Visual Docking Framework}
\label{section2}

\begin{figure*}[t]
    \centering
    \includegraphics[width=1\linewidth]{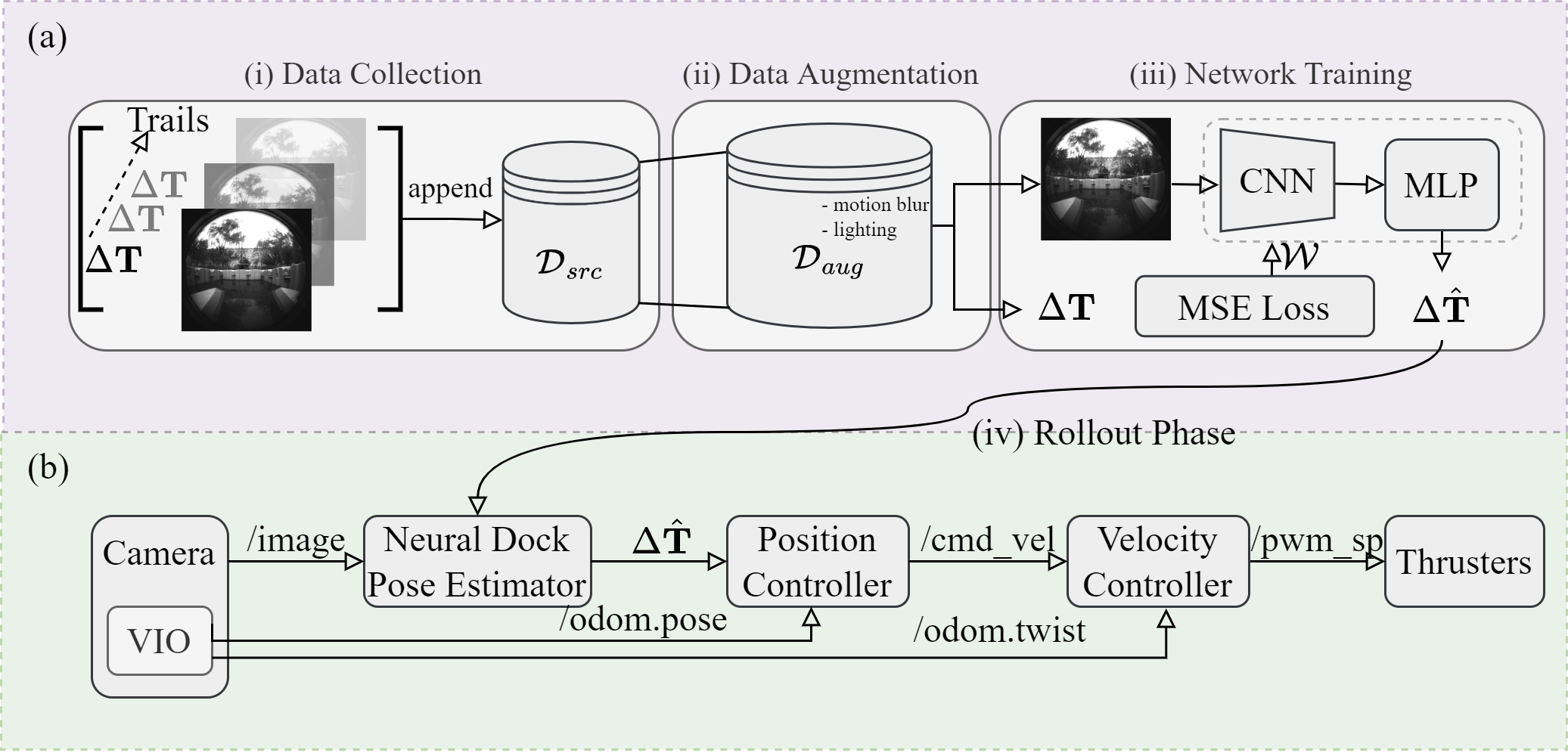}
    \caption{Overview of the framework. (a) The pipeline of data collection, augmentation and NPDE training. (b) The visual-servo architecture for PBVS-based autonomous docking task.}
    \label{fig:lvd-overview}
\end{figure*}

The Supervised Learning-based Visual Docking Framework is an autonomous visual docking system using supervised learning on a sequence of fisheye images with auto-labeling. As summarized in Figure \ref{fig:lvd-overview}, the pipeline for our framework involves five phases: (a.i) data collection and auto-labeling, (a.ii) dataset augmentation, (a.iii) model training, (a.iv) neural dock pose estimation, and (b) motion controller.

\subsection{Coordinates}

\begin{figure}[htbp]
    \centering
    \includegraphics[width=1\linewidth]{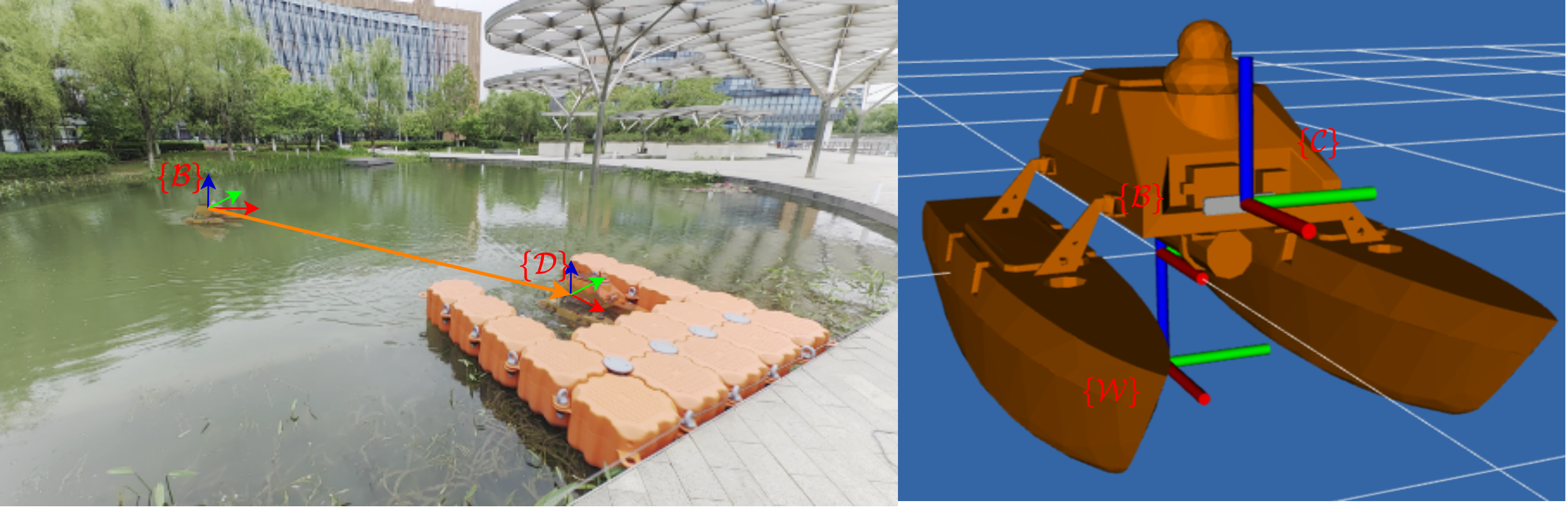}
    \caption{Illustration of the coordinate frames.}
    \label{fig:reference-frame}
\end{figure}

Figure \ref{fig:reference-frame} indicates the coordinate frames of the USV and the dock. The dock frame $\{\mathcal{D}\}$, defines the desired docking frame of the USV; the base frame of the USV $\{\mathcal{B}\}$, defines the current body frame of the USV; the world frame $\{\mathcal{W}\}$, defines the fixed world frame; the camera frame $\{\mathcal{C}\}$ is fixed relative to the $\{\mathcal{B}\}$; The dock frame $\{\mathcal{D}\}$ is a dummy coordinate, however, it can be represented by the docked point where the user deploys the USV initially at the dock. For arbitrary two frames $\{\mathcal{I}\}$ and $\{\mathcal{J}\}$,   $^{\{\mathcal{J}\}}\mathbf{T}_{\{\mathcal{I}\}} \in \mathbf{SE}(3)$ is a homogeneous transformation matrix representing the pose of $\{\mathcal{I}\}$ frame in $\{\mathcal{J}\}$ frame, which is denoted by Eqn. \ref{eqn:tf}:

\begin{equation}
\label{eqn:tf}
{ }^{\{\mathcal{J}\}} \mathrm{T}_{\{\mathcal{I}\}}=\left[\begin{array}{cc}
{ }^{\{\mathcal{J}\}} \mathrm{R}_{\{\mathcal{I}\}} & { }^{\{\mathcal{J}\}}p_\mathrm{org} \\
0 & 1
\end{array}\right]
\end{equation}
where ${ }^{\{\mathcal{J}\}} \mathrm{R}_{\{\mathcal{I}\}}$ is a rotation matrix, and ${ }^{\{\mathcal{J}\}} p_\mathrm{org}$ is the coordinates of the origin of the coordinate system $\{\mathcal{I}\}$ in the coordinate frame $\{\mathcal{J}\}$. Then, the relative transformation between the dock and the USV is denoted as Eqn. \ref{eqn:deltaT}:

\begin{equation}
\label{eqn:deltaT}
    \Delta \mathbf{T} := ^{\{\mathcal{B}\}}\mathbf{T}_{\{\mathcal{D}\}} = ^{\{\mathcal{B}\}}\mathbf{T}_{\{\mathcal{W}\}} \ ^{\{\mathcal{W}\}}\mathbf{T}_{\{\mathcal{D}\}}
\end{equation}
where $^{\{\mathcal{B}\}}\mathbf{T}_{\{\mathcal{W}\}}$ is calculated by the Visual-Inertial Odometry (VIO) module of the USV, and $^{\{\mathcal{W}\}}\mathbf{T}_{\{\mathcal{D}\}}$ is the desired dock pose defined by users. $\Delta \mathbf{T}$ represents the relative pose between the USV base frame and the dock frame. We also assume that the roll and pitch angles of the USV are relatively small or in a stable period pattern, and the dock and the USV both float at sea level. Therefore, the $\Delta \mathbf{T} \in \mathbf{SE}(3) $ can be reduced to $\Delta \mathbf{T} \in \mathbf{SE}(2)$, where the state variables are $x$, $y$ and $\theta$ for the position in the world frame and the yaw angle of the USV, respectively. Through this design, our data collection method achieves the auto-labeling feature of the images with relative pose between the dock and the USV.

\subsection{Data Collection and Augmentation}

\subsubsection{Dataset Collection} 

The process begins with the initialization of a source dataset \(\mathcal{D}_{\text{src}}\) for the autonomous docking task. For each new scene, the dock is randomly placed in the water environment, followed by positioning the USV at the dock pose at the initial docked pose. The camera mounted on the USV captures a first-person-view fisheye image, denoted as \(\mathbf{I}\). At timestep \(t_i\), \((\mathbf{I},\Delta \mathbf{T})_{t_i}\) represents a datapoint, which is the \(\mathbf{I}\)-\(\Delta \mathbf{T}\) pair at that timestep. Subsequently, the USV moves randomly around the pre-docking area \(\mathcal{S}\) through teleoperation to add the \(\mathbf{I}\)-\(\Delta \mathbf{T}\) pairs to the source dataset \(\mathcal{D}_{\text{src}}\). Here, \(\Delta \mathbf{T}\) serves as the label for supervised learning, encoding the relative pose information between the USV and the dock. It is obtained by the current pose from VIO and the initial pose of the dock, which is fixed in each data collection trail. The dock pose is fixed in a data collection scene. 

Data synchronization is achieved through \textit{message\_filters} ROS package and \textit{ApproximateTime} function which allows for coordinating messages from different ROS nodes that may not have perfectly synchronized clocks. ApproximateTime is achieved by timestamping messages with their respective ROS Time and then using a buffer to match messages that are close enough in time within a specified threshold. When messages are published with timestamps, they are inserted into a synchronized buffer. Subscribers then request messages with a desired timestamp, and the buffer returns the message with the closest timestamp within the specified threshold. By synchronizing data in this way, our implementation enables more accurate and reliable data logging.

\begin{figure}[t]
    \centering
	  \subfigure[]{
       \includegraphics[width=0.3\linewidth]{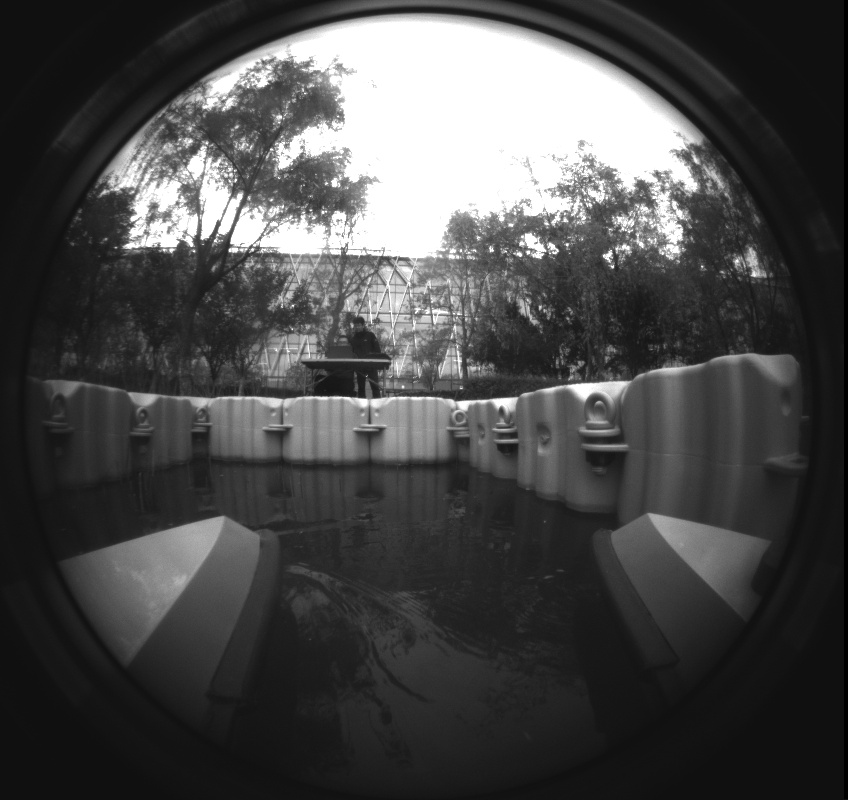}}
    \label{A1}\hfill
	  \subfigure[]{
        \includegraphics[width=0.3\linewidth]{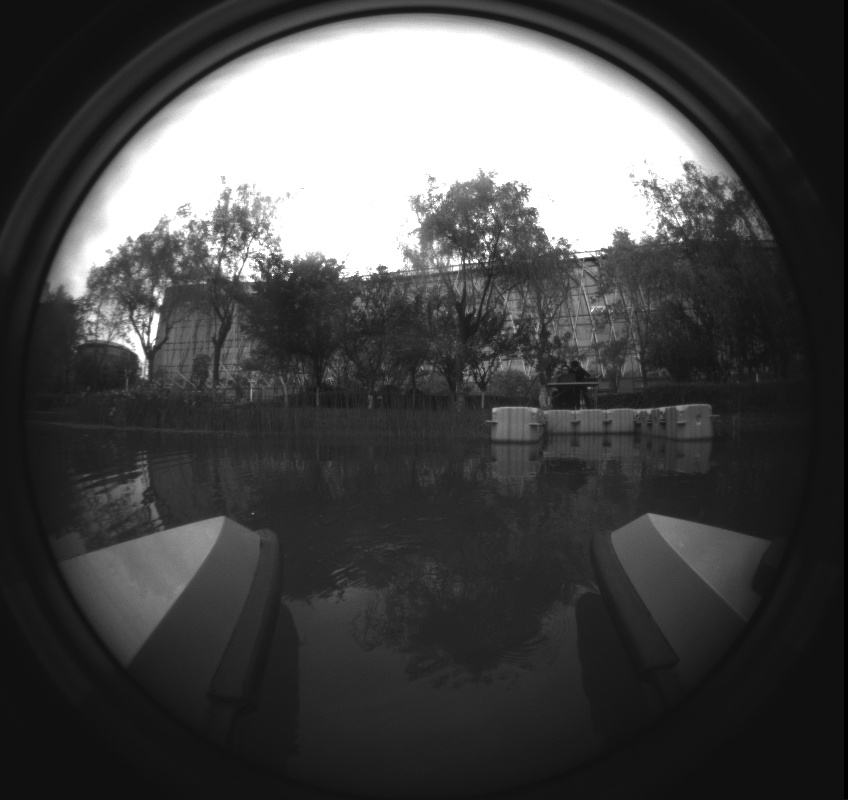}}
    \label{B1}\hfill
	  \subfigure[]{
        \includegraphics[width=0.3\linewidth]{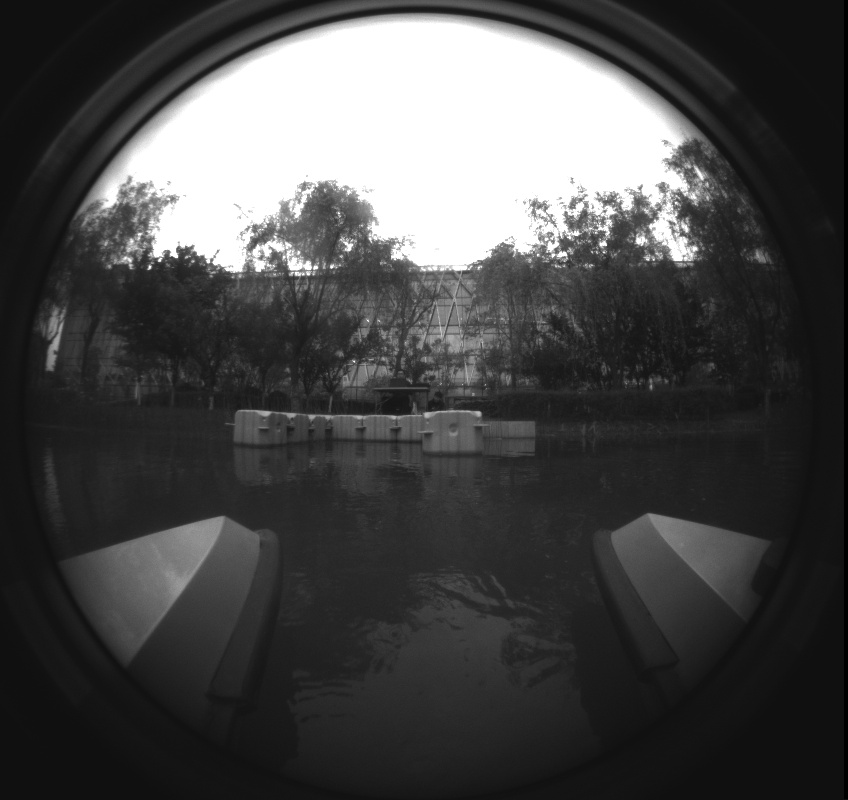}}
    \label{C1}\hfill
	  \subfigure[]{
       \includegraphics[width=0.3\linewidth]{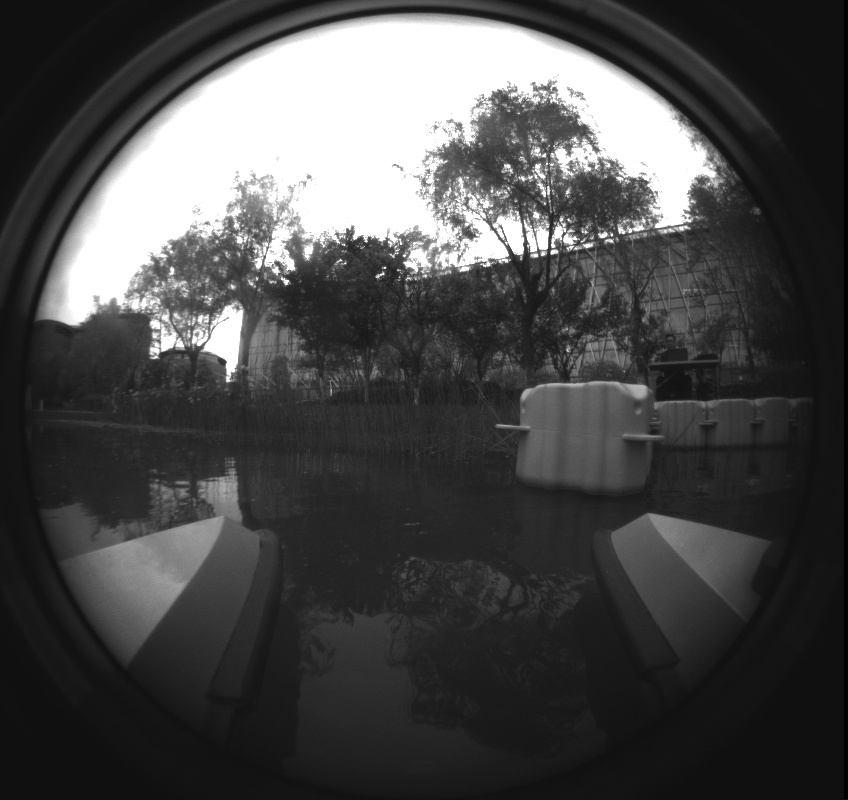}}
    \label{D1}\hfill
	  \subfigure[]{
        \includegraphics[width=0.3\linewidth]{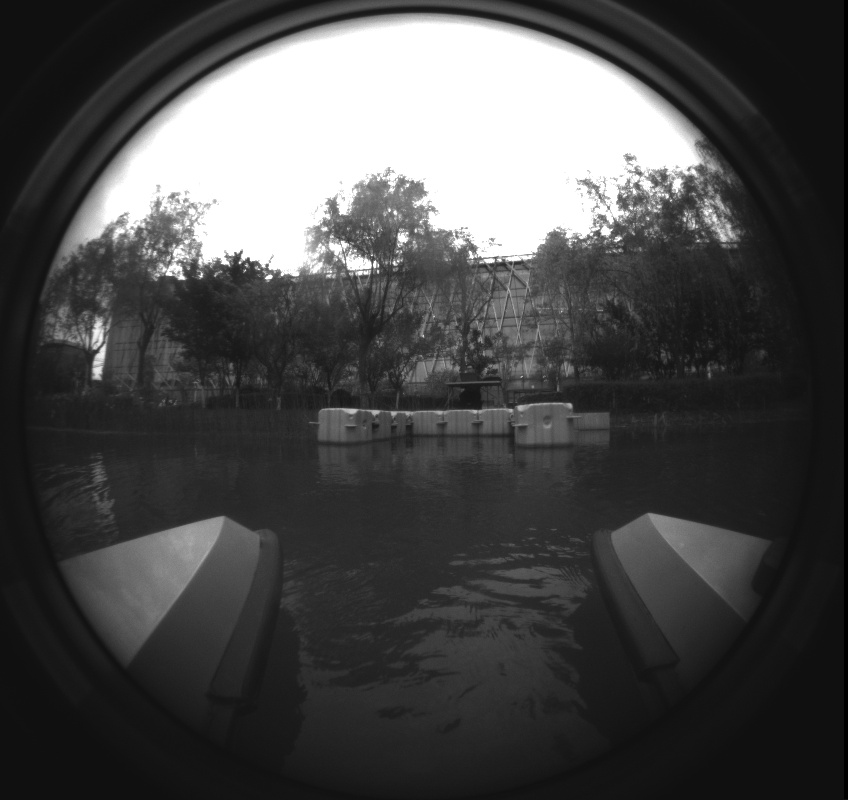}}
    \label{E1}\hfill
	  \subfigure[]{
        \includegraphics[width=0.3\linewidth]{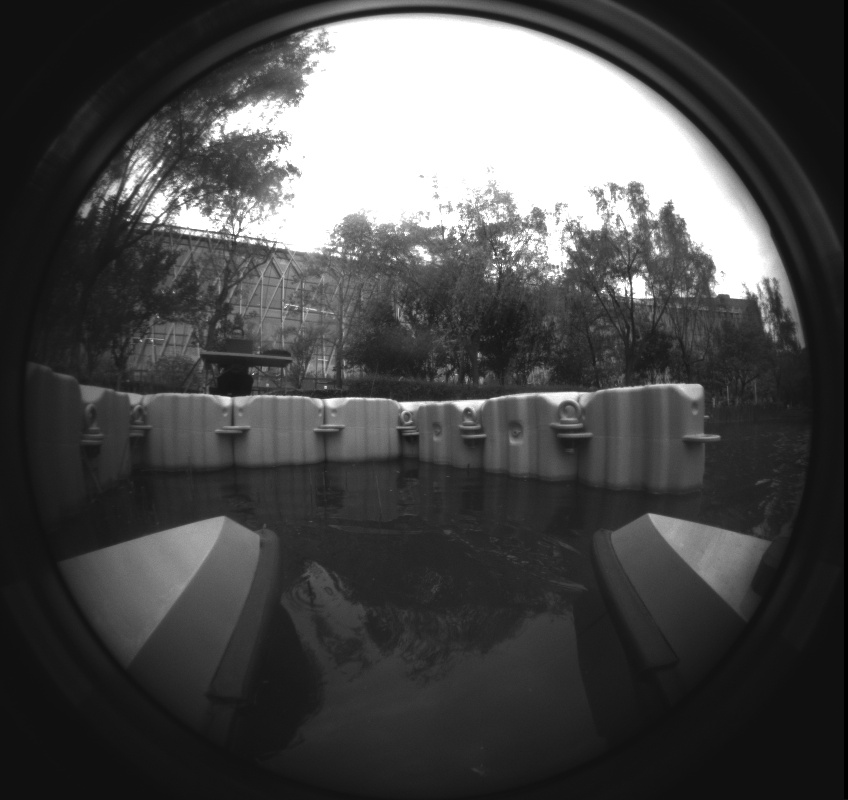}}
    \label{F1}\hfill
	  \caption{Samples of logged data pairs.}
	  \label{fig:sample1} 
\end{figure}

Our data collection pipeline design offers several benefits. It eliminates the need for humans to manually label data. Human teleoperation is only required when resetting the USV to the dock pose or in safe-critical situations. Additionally, the framework does not require camera calibration for pose regression since the camera is rigidly mounted on the USV. The camera's intrinsic and extrinsic parameters are implicitly represented in the fisheye image. Figure \ref{fig:sample1} demonstrated six samples of fisheye images at different poses related to the dock.  

\subsubsection{Dataset Augmentation}

\begin{figure}[p]
    \centering
	  \subfigure[]{
       \includegraphics[width=0.3\linewidth]{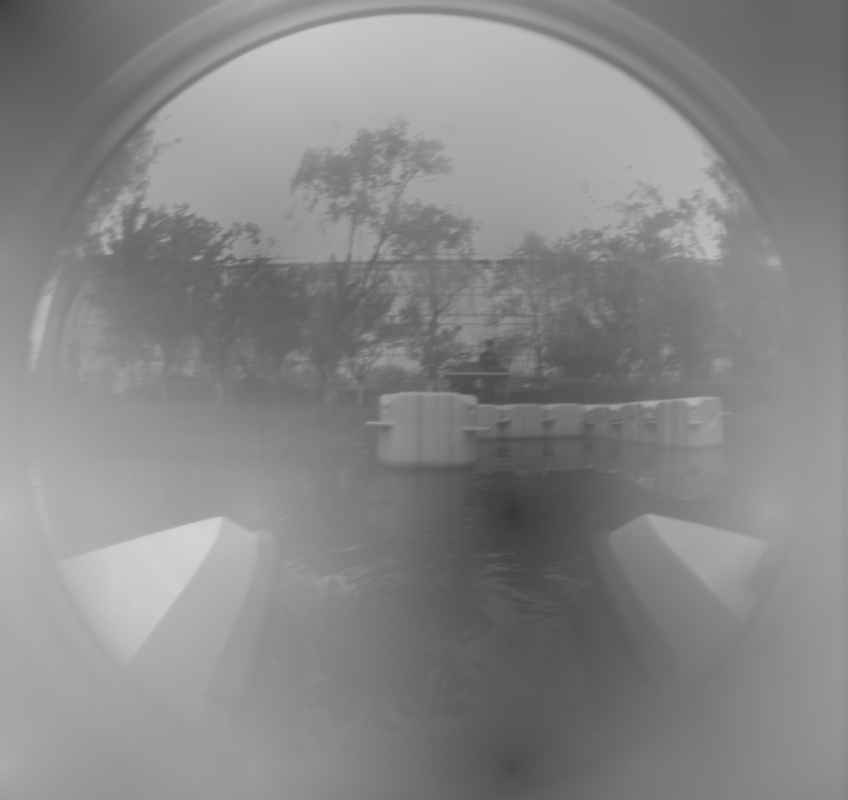}}
    \label{A2}\hfill
	  \subfigure[]{
        \includegraphics[width=0.3\linewidth]{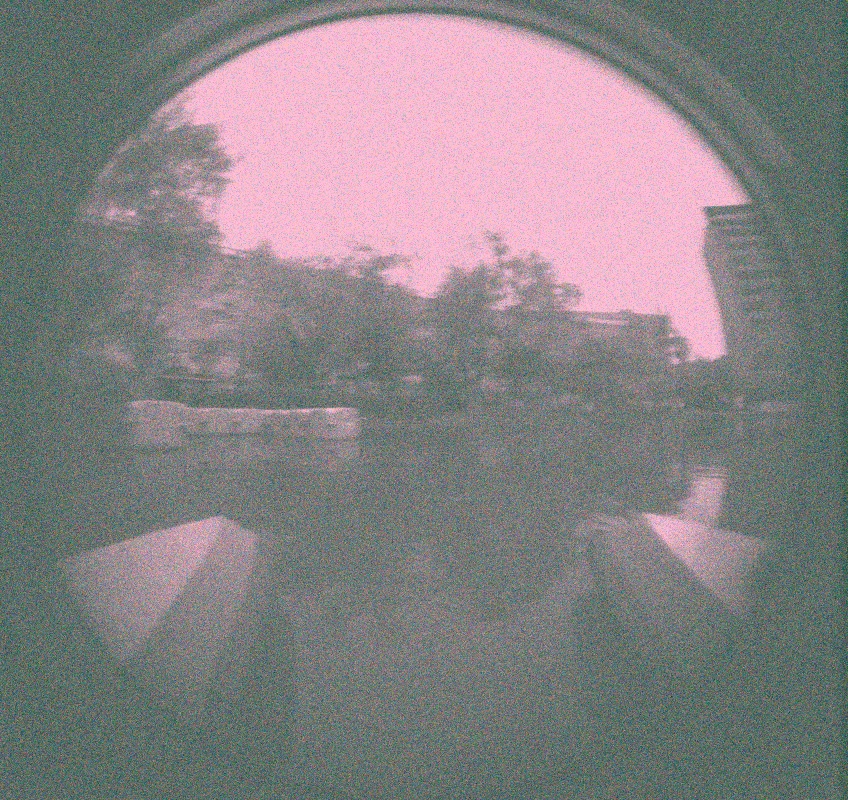}}
    \label{B2}\hfill
	  \subfigure[]{
        \includegraphics[width=0.3\linewidth]{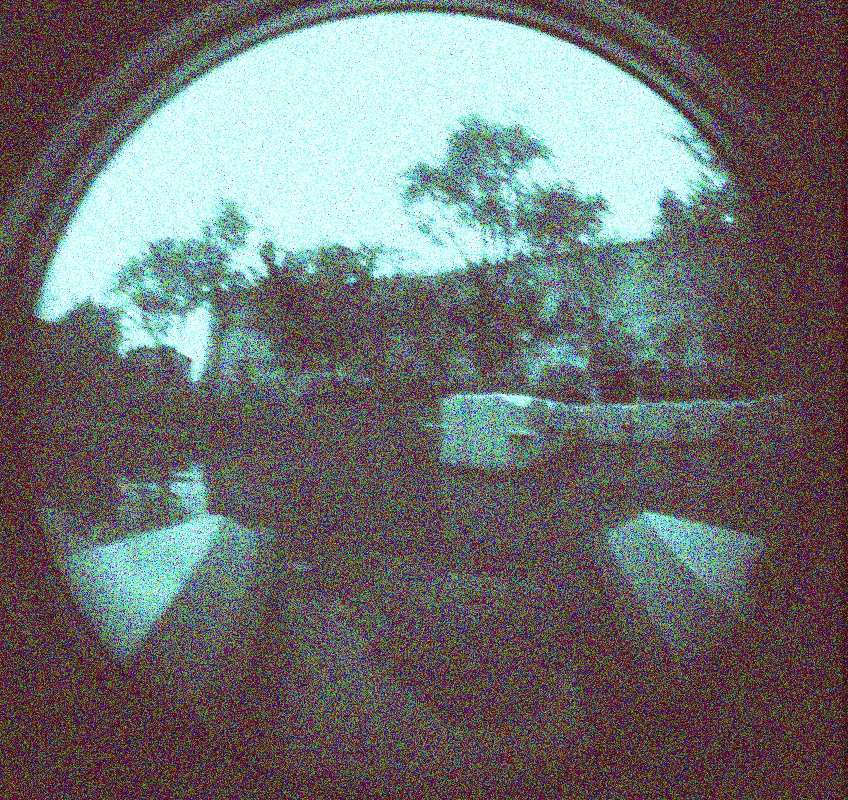}}
    \label{C2}\hfill
	  \subfigure[]{
       \includegraphics[width=0.3\linewidth]{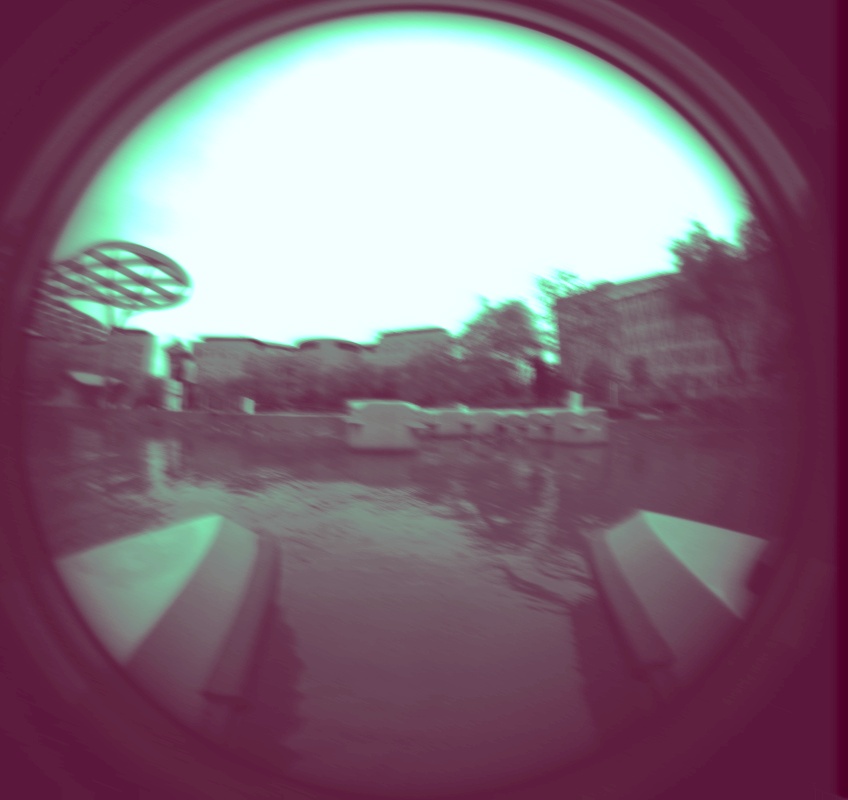}}
    \label{D2}\hfill
	  \subfigure[]{
        \includegraphics[width=0.3\linewidth]{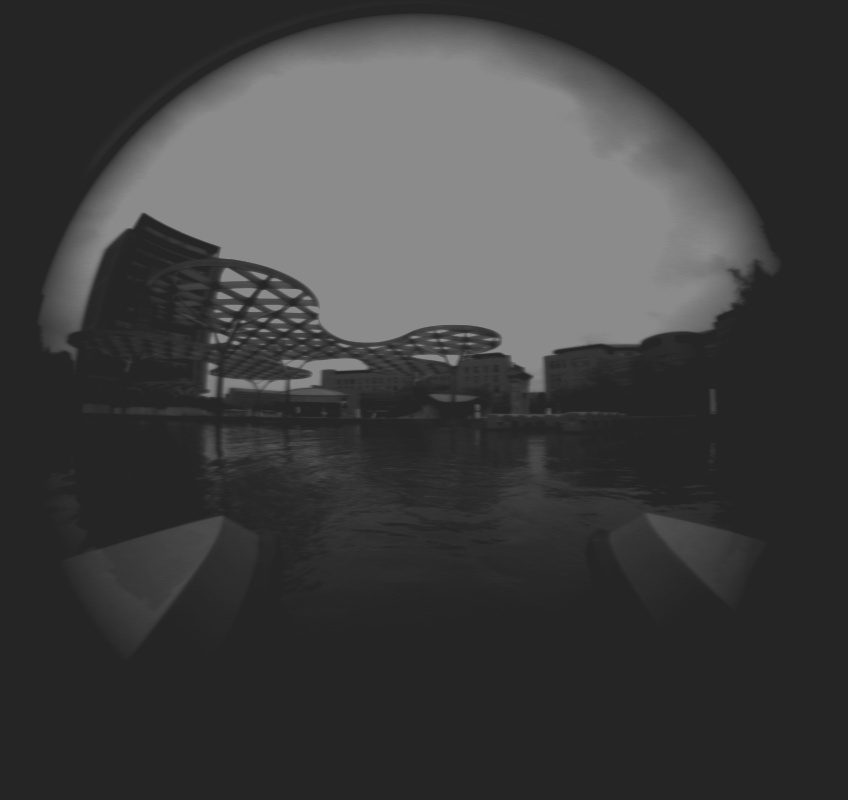}}
    \label{E2}\hfill
	  \subfigure[]{
        \includegraphics[width=0.3\linewidth]{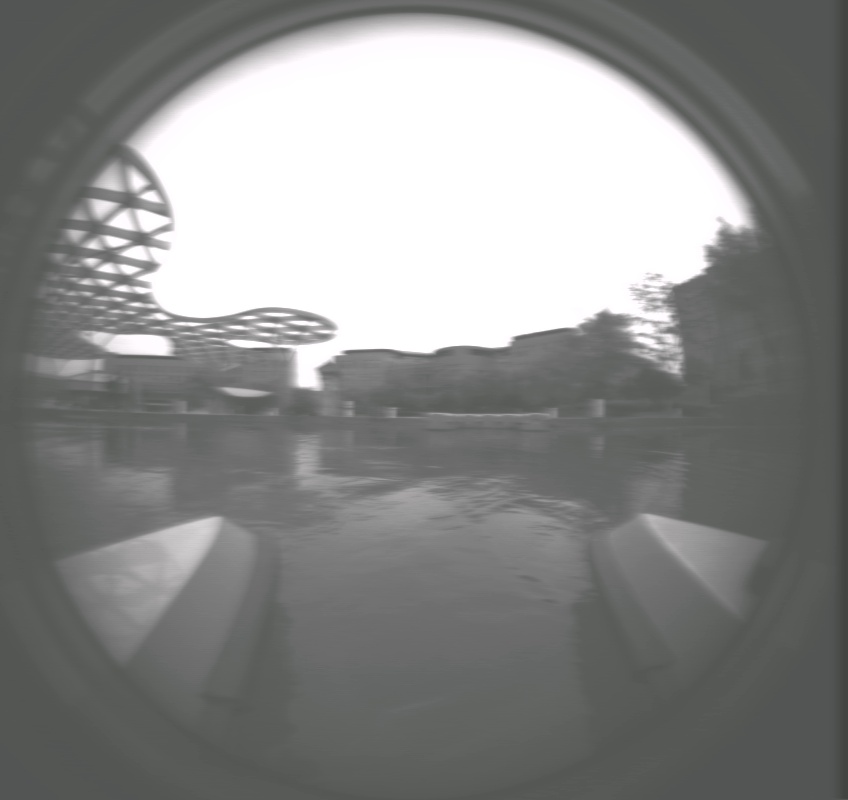}}
    \label{F2}\hfill
	  \caption{Dataset augmentation in two experimental scenarios.}
	  \label{fig:augmentation} 
\end{figure}

The diverse source dataset $\mathcal{D}_{\text{src}}$ is created by introducing variations in water environments and different initial states. To further enhance the dataset, we apply a series of image operations that reflect to the complexity and dynamics of water environments. These operations include adding Gaussian noise and dropping pixels to augment data in the noise domain, as well as adding motion blur to simulate images captured by fast-moving USVs. Additionally, we vary the saturation and brightness of the water and dock elements and overlay weather effects such as fog and rain onto the images. The augmentation process uses the library \textit{imgaug} \cite{imgaug}. For details on the collection and randomization methods, as well as samples of the dataset, please refer to section \ref{subsub:data-sampl}. Figure \ref{fig:augmentation} illustrates dataset augmentation in two experimental scenarios. The top row (a, b, c) represents the augmented images from the first experimental scenario, while the bottom row (d, e, f) showcases the results from the second scenario.

\subsection{Neural Dock Pose Estimator}

\subsubsection{Model Structure}
\label{method:train}

The Neural Dock Pose Estimator (NDPE) is trained using supervised learning with a neural network, denoted as $f$, to predict the relative pose $\Delta \hat{\mathbf{T}}$ based on input images $\mathbf{I}$. This is represented as $f: \mathbf{I} \rightarrow \Delta \hat{\mathbf{T}}$, where $\hat{}$ indicates estimated values. Our implementation uses a convolutional neural network (CNN) with the PyTorch framework based on a VGG-19 backbone. The CNN's convolutional layers extract features from images, which are then used by a multi-layer neural network to predict the relative pose. The CNN architecture consists of a bunch of convolutional layers with ReLU activation and max-pooling for downsampling, followed by three fully connected layers with ReLU activation and Dropout. Input images are cropped and resized to a 224 x 224 monochrome images using OpenCV, and the loss function is based on MSE. The relative pose labels are normalized when loading data from a dataloader. Stochastic Gradient Descent (SGD) is used as an optimizer. The dataset is split randomly into 80\% for training and 20\% for validation. The model is trained for 100 epochs with a batch size of 32, and the best model's weights $\mathcal{W}$, which achieved the lowest validation loss for the validation set, are saved as a weight file for pose prediction. Note that the model training is based on the augmented dataset, so the test dataset includes scenarios with different generated motion blur, different lighting conditions, and so on. If the model can perform well on the generated data, it shows that our method is very effective. Moreover, the weights of the CNN were not frozen. The entire NDPE was jointly fine-tuned based on our dataset.

\subsubsection{Loss Function}

The Mean Squared Error (MSE) loss function is a commonly used measure in machine learning and regression problems to quantify the difference between predicted and actual values. Mathematically, the MSE is calculated as the average of the squared differences between the predicted values $\Delta\hat{\mathbf{T}}$ and the actual values $\Delta \mathbf{T}$:

\begin{equation}
\mathrm{MSE}=\frac{1}{m} \sum_{i=1}^m\left(\Delta \mathbf{T}_i-\Delta \hat{\mathbf{T}_i}\right)^2
\end{equation}

\noindent where $m$ represents the batch size, and $i$ represents the index of each predicions.

\subsubsection{Model Rollout}

During the rollout phase, the trained weight for pose estimation is loaded. For each rollout, an image $\mathbf{I}$ and the robot's current pose $^{\{\mathcal{B}\}}\mathbf{T}_{\{\mathcal{W}\}}$ are captured from VIO. The model uses the image to predict the estimated relative pose $\Delta \hat{\mathbf{T}}$. The dock pose $^{\{\mathcal{W}\}}\mathbf{T}_{\{\mathcal{D}\}}$, where the USV should move to, is calculated.  

Some readers may confused that our system already has a global dock position. So the USV can just move directly to the position. The position of the USV at the very beginning of its journey is known. Then with the VIO we can also constantly get the position of the USV in real time. But this position may have some drift. Please note that our target position is constantly updated, not fixed, and the NDPE module will estimate the latest relative position based on the image, which is added to the current position to update the final target dock position. The docking error is reduced by the closed-loop visual servo. We will introduce more implementation details of our model in Section \ref{sec:experimentsandresults}.

\subsubsection{Visual Servo}

We incorporate a PBVS implementation for the USV to track the desired dock pose, enabling the USV to navigate to the dock pose using a functional motion controller. It is worth emphasizing that our VS framework is position-based rather than image-based. NDPE has estimated the relative position of the dock from the image. The goal of the position servo is the latest position updated online. The accuracy of the estimation is assumed to be highest when the USV is near the dock. In the first single-frame implementation, the USV can navigate to the desired dock pose using a single-frame rollout, marking the initial implementation of our framework. However, accuracy in sequential estimation is assumed to be highest as the USV approaches the dock. In the last single-frame implementation, we performed rollouts and updated the desired dock pose at each timestep, allowing the pose estimation error to decrease over time. This represents the final single-frame implementation of our framework for high-rate closed-loop control. Evaluation can be found in section \ref{subsec:vs-exp}.

\subsection{Motion Controller}

\subsubsection{PID Controller}

A Proportional-Integral-Derivative (PID) controller is employed to control the USV. The outer controller takes the pose error, which is the error between the current USV pose and the estimated dock pose, as input. It then produces the desired velocity as output. The inner controller's input is the velocity error, which is the difference between the current USV velocity and the desired velocity. A PID controller calculates an error signal $e(t)$ iteratively by comparing the reference signal $r(t)$ with a measurement $y(t)$, and then computes a feedback signal $u(t)$ by summing over proportional, integral, and derivative terms. The implementation of the motion controller in ROS is adapted from Samak et al \cite{noauthor_tinker-twinssingaboat-vrx_nodate}.

\subsubsection{Control Allocation}

To deal with differential thrust using a fixed configuration of two direct drive thrusters, the equations of motion for the USV are often simplified and expressed in terms of thrust and scaling coefficients. The simplified equations are given by:

\begin{align}
v_{usv}&=\alpha \left(T_r+T_{\ell}\right) \label{eqn:model1.1} \\
\omega_{usv}&=\beta \left(T_r-T_{\ell}\right)
\label{eqn:model1.2}
\end{align}

\noindent where $T_r$ and $T_{\ell}$ represent the proportional thrust of the right and left engines, respectively, while $\alpha$ and $\beta$ are scaling coefficients based on the USV's size and the distance between the two engines. However, for Eqns. \ref{eqn:model1.1} and \ref{eqn:model1.2} can be challenging to interpret in terms of how quickly the USV is moving horizontally and rotating alog z-axis. To address this, we can express the motion of the USV in terms of wheel velocities using variables $v$ and $\omega$, which intuitively allocate motion control. The inverse kinematics of the USV can be represented as:

\begin{align}
    T_r&=\frac{\beta v_{usv} + \alpha \omega_{usv}}{2 \alpha \beta} \\
    T_{\ell}&=\frac{\beta v_{usv} - \alpha \omega_{usv}}{2 \alpha \beta}
\end{align}

The motion control mixer then maps the linear velocity and angular rate of the USV to proportional thrusts based on our USV configuration. We also assume that the thrust command is proportional to the pulse-width modulation (PWM) setpoint. Then define the transformation of the PWM setpoints for the actuators as:

\begin{align}
    {pwm}_{r} &=\frac{\beta u_{v} + \alpha u_{\omega}}{2 \alpha \beta}\\
    {pwm}_{\ell} &=\frac{\beta u_{v} - \alpha u_{\omega}}{2 \alpha \beta}
\end{align}

To ensure that the PWM value falls within a valid and safe range for the system, a PWM limiter is employed using the CLIP function:

\begin{align}
    pwm_{out,i}=\text{CLIP}(pwm_{i},pwm_{min},pwm_{max})
\end{align}

\noindent where $i=\{r, \ell\}$ and $pwm_{max}$ and $pwm_{min}$ represent the maximum and minimum valid PWM values, respectively.

\section{Real-world Experiments and Results}
\label{sec:experimentsandresults}

\subsection{Hardware Implementation}

\begin{figure}[!ht]
    \centering
	  \subfigure[The dock.]{
       \includegraphics[width=0.47\linewidth]{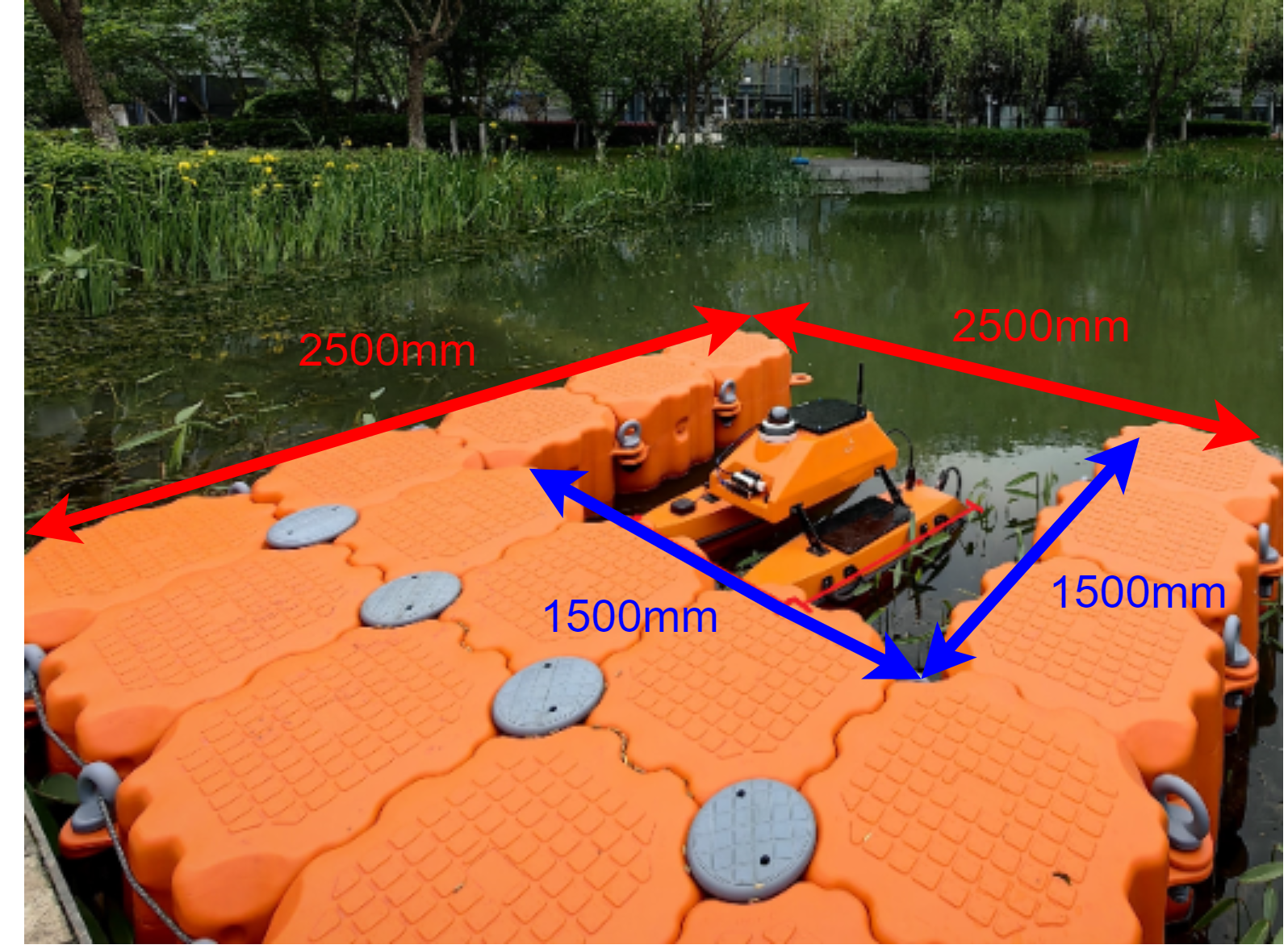}}
    \label{fig:hardwareaa}\hfill
	  \subfigure[The USV.]{
        \includegraphics[width=0.47\linewidth]{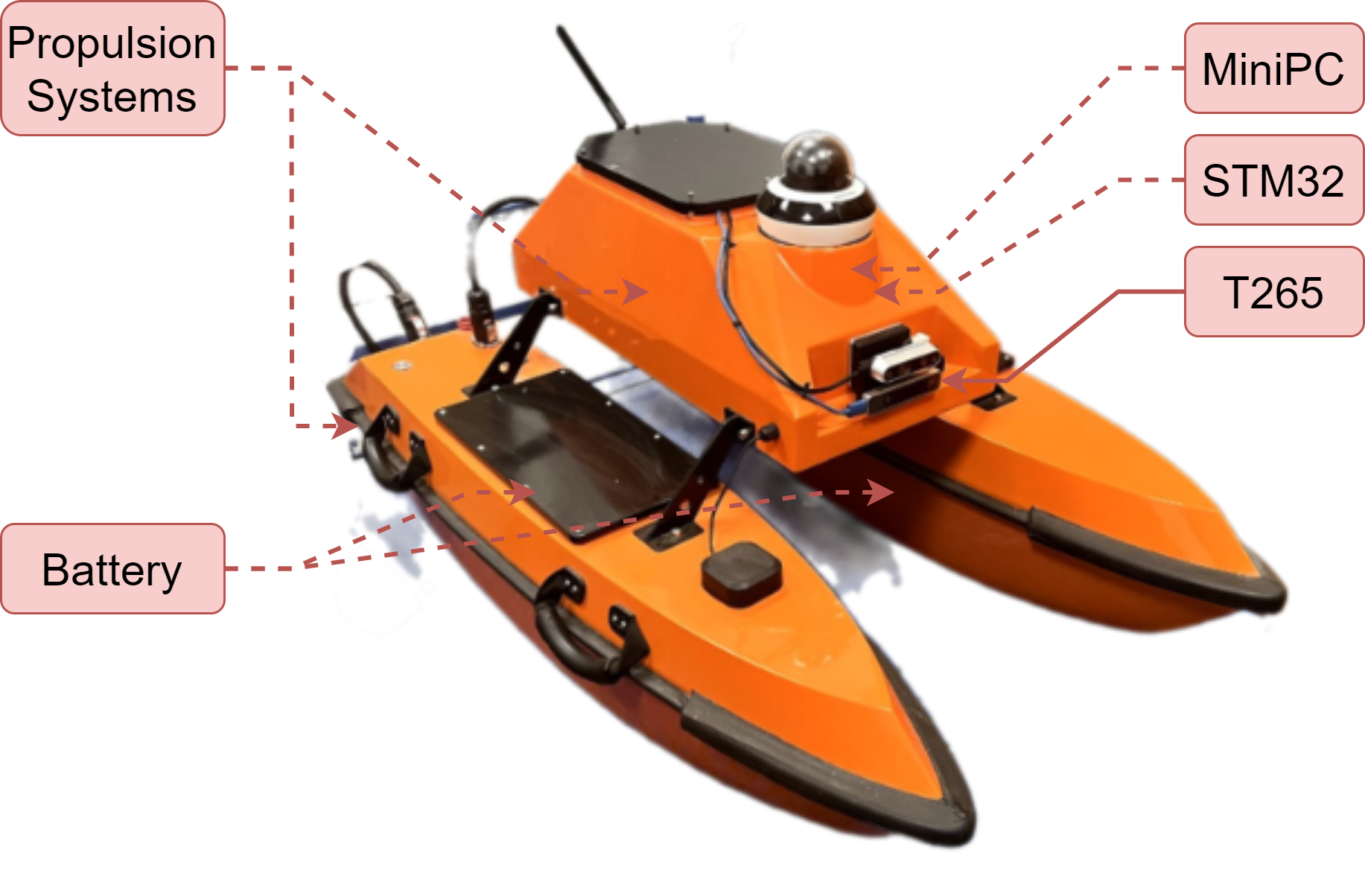}}
    \label{fig:hardware_b}\hfill\\
	  \subfigure[Hardware topology of the USV.]{
        \includegraphics[width=0.9\linewidth]{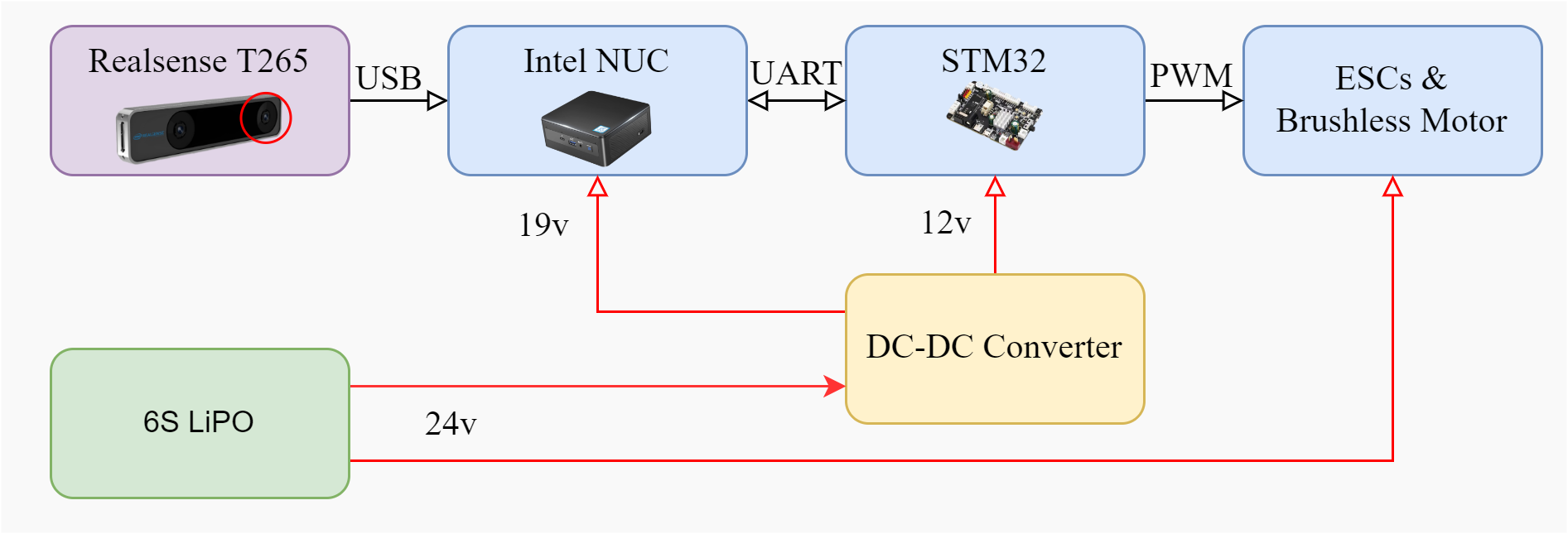}}
    \label{fig:hardware_c}\hfill\\
	  \caption{Descriptions of the hardware and components' connection.}
	  \label{fig:hardware} 
\end{figure}

We use 16 floating blocks to setup a U-shaped dock with a 1500 x 1500 mm docking area for our USV, shown in Figure \ref{fig:hardware} (a). The propulsion systems of the USV with two durable thrusters at the rear end of the USV, as demonstrated in Figure \ref{fig:hardware} (b). The mechanical dimensions of the USV are 1100 x 780 mm. The material of the USV structure is carbon fiber composite and CNC parts. We also add a T265 camera on the front of the USV using a 3D-printed mount. More detailed properties of the USV can be found in Table~\ref{tab:env_param}.

The illustration of the hardware topology block diagram of the USV is shown in Figure \ref{fig:hardware} (c). The Intel Realsense Tracking Camera T265 provides Visual Inertial Odometry (VIO) information and fisheye images that can be used for visual docking. The diagonal FOV of a fisheye image is 173 degrees. Therefore, it ensures the high visibility of the dock when there is a large yaw shift. The size of the image is 848 x 800 pixels. The feature of the global shutter may reduce the motion blur and pixel displacement. The camera is refreshed as 30 FPS providing monochrome images. T265 also provides 6-DoF odometry data at a sample rate of 200Hz, so we obtain the state estimation for control from it. It is worth mentioning that we initially used D435 RGB camera to obtain images. However, its horizontal FOV is 69 degrees which the USV was not able to have persistent observation of the dock. The camera connects with Intel NUC 10i7FNH through a USB 3.0 cable. Intel NUC performs high-level computations with ROS. It runs Ubuntu 20.04 and ROS1 Noetic. STM32 provides low-level control for the thrust systems, which parse the desired PWM setpoints to the ESCs through PWM peripherals. The communication between NUC and STM32 is through UART with customized protocol. For software architecture, please see the previous section \ref{section2}. Table~\ref{tab:env_param} also summarizes the components and their specifications of our USV systems.

\begin{table}[t]
\caption{Hardware specifications.}
\label{tab:env_param}
\centering
\begin{tabular}{@{}lllll@{}}
\toprule
\textbf{Component/Parameter}        & \textbf{Specification/Properties}   \\ \midrule
Onboard Computer          & Intel NUC 10i7FNH           \\
Camera \& VIO             & RealSense T265\\
Embedded Controller       & STM32F103RCT6               \\
Battery                   & 6S1P LiPO 55C               \\
DC-DC Converter           & 19V/10A, 12V/5A             \\
ESC                       &  2S-6S/90A/bidirectional    \\
Brushless Motors          &  Direct-drive motor 300KV   \\
Blades                    & 8-cm three-bladed paddle    \\
WiFi adapter              & Intel Wireless-AC 9462      \\ \midrule
Dock Size  [L/W]        & 2500x2500 mm                \\
Docking Area Size  [L/W]  & 1500x1500 mm                 \\
USV Frame Size [L/W]      & 1100x780 mm                 \\ \bottomrule
\end{tabular}
\end{table}

\newpage

\subsection{Data Collection Procedures}
\label{subsub:data-sampl}

\begin{figure}[htb]
    \centering
	  \subfigure[Distribution of distance between the dock and the USV.]{
       \includegraphics[width=0.45\linewidth]{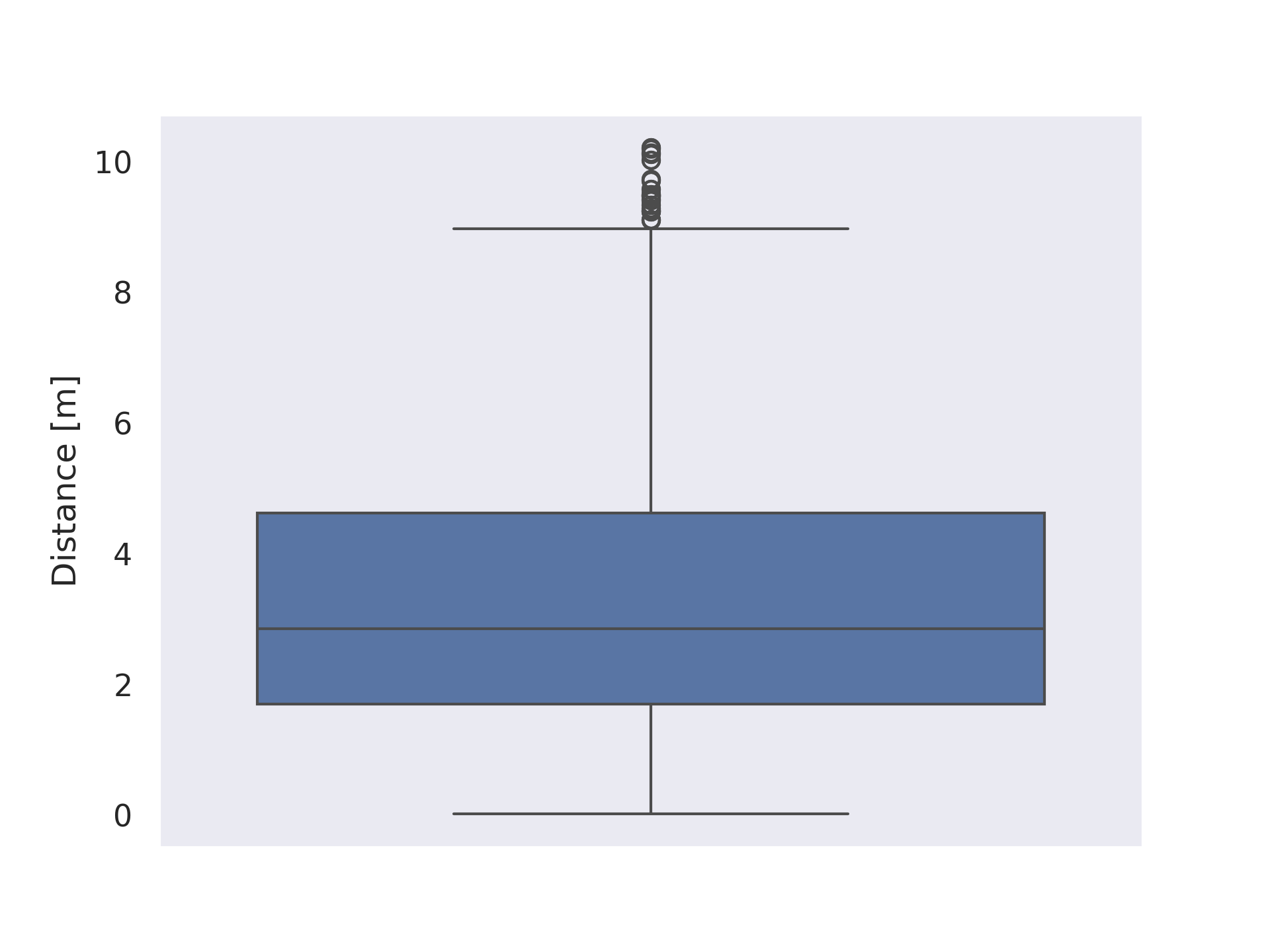}}
    \label{fig:dataset_distribution1}\hfill
	  \subfigure[Distribution of the velocity profile of the USV.]{
        \includegraphics[width=0.45\linewidth]{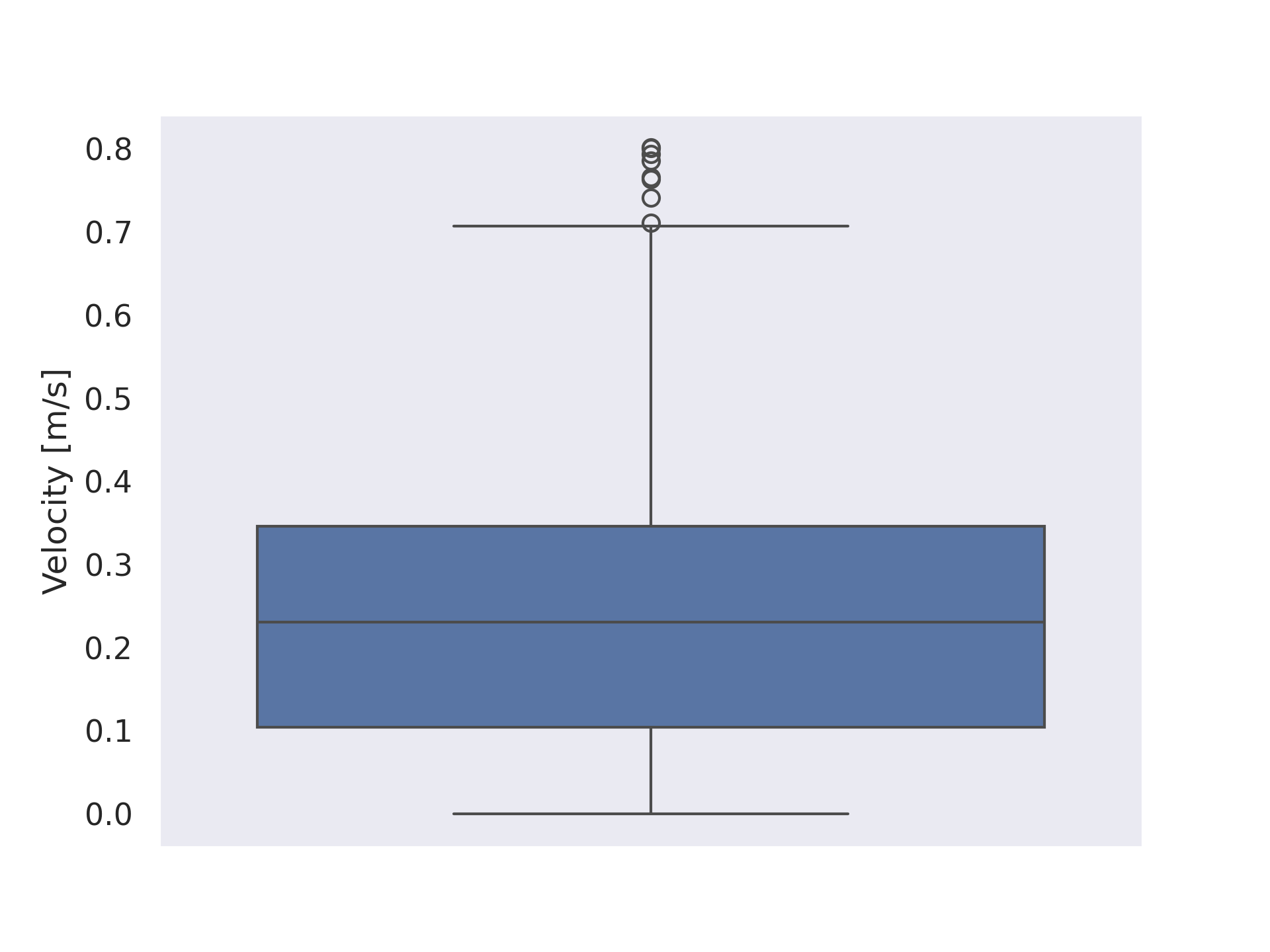}}
    \label{fig:dataset_distribution2}\hfill
	  \caption{Distributions of the dataset.}
	  \label{fig:dataset_distribution} 
\end{figure}

We conducted multiple field trips to collect real-world data across different scenarios. Firstly, we initially deployed the USV at the docked position for each instance. Secondly, we powered on all systems and connected to the NUC through a remote desktop connection. Next, the USV randomly moves within the pre-docking area far away from the dock. We ensured the dock was visible during the collection process. A synchronized data logger continuously appended the $\mathbf{I}$-$\Delta \mathbf{T}$ data to the dataset. In total, we collected 2K data pairs in water environments. As shown in Figure \ref{fig:dataset_distribution}, the distribution of distance between the dock and the USV, and the distribution of the velocity profile of the USV are visualized using boxplots. For Figure \ref{fig:dataset_distribution} (a), the average distance within the dataset is 3.29 m with a standard deviation of 2.01 m. Additionally, the average velocity of the dataset is 0.24 m/s with a standard deviation of 0.16 m/s in Figure \ref{fig:dataset_distribution} (b). The augmented dataset is generated based on the source data and has the same data distribution. The only difference is that there are perturbations on the images using the previously described methods to simulate more difficult-to-capture scenarios such as motion blur, different lighting conditions, etc.

\subsection{Experiments on NDPE}
\label{subsub:sdpe-res}

The NDPE network was trained based on our gathered data. We performed model training on a laptop (Intel i7-12700H processor and NVIDIA GeForce RTX 3070 Laptop GPU with CUDA). 80\% of data points were used for model training, and 20\% for model validation. The batch size was set at 32. A Stochastic Gradient Descent (SGD) optimizer with 0.001 learning rate and 0.9 of a momentum factor was used for the learning optimization. The loss function is based on MSE calculated by batch labels and predictions. The model was trained over 100 epochs and took about 1.5 hours. We saved the best model weight based on the minimum validation loss. More details about the model implementation can be found at Sec. \ref{method:train}.

\begin{figure}[htb]
    \centering
       \includegraphics[width=1\linewidth]{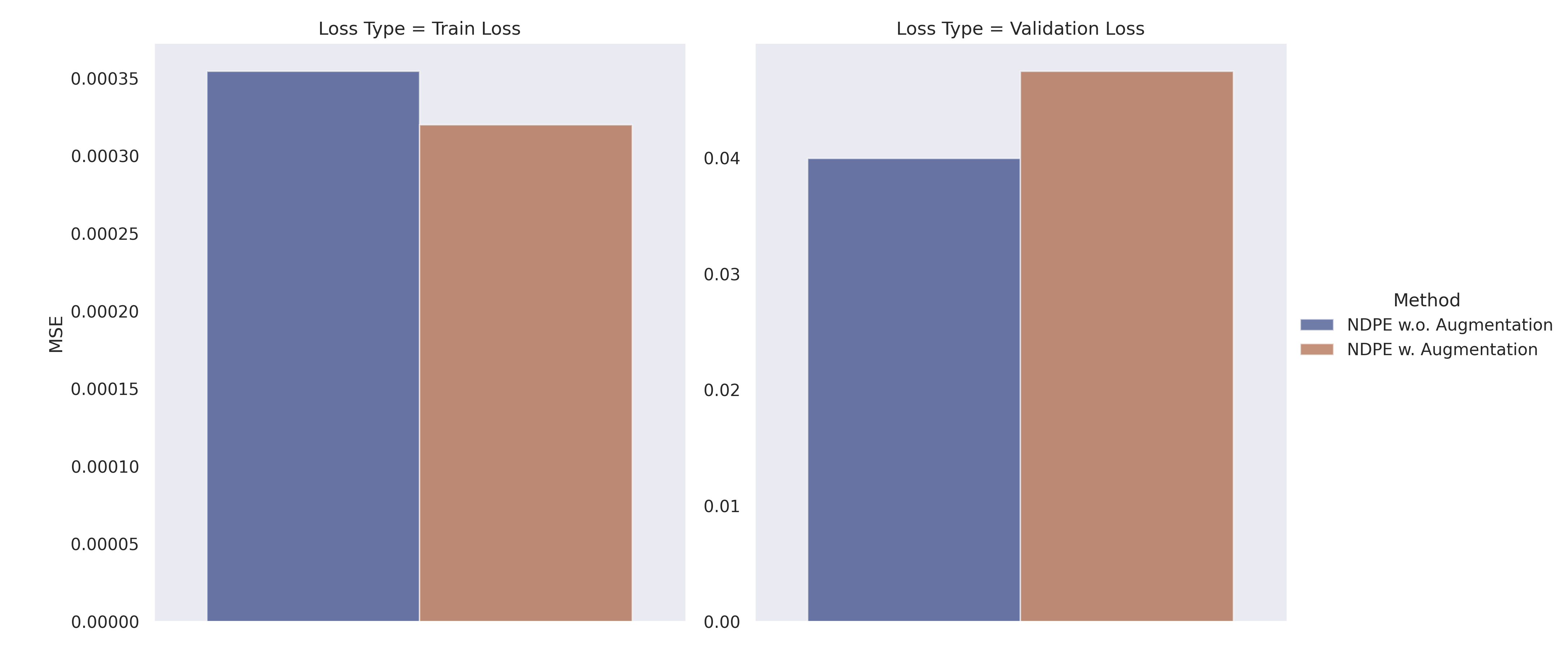}
    \label{fig:trans1}\hfill
	  \caption{The NDPE training loss.}
	  \label{fig:res1} 
\end{figure}

After completing the training and validation phases, the model demonstrated the capability of high accuracy for dock pose estimation. In Figure \ref{fig:res1}, the training loss converged significantly, reaching a minimum value of 0.00035 m and 0.00030 m for NDPE without and with augmentation, respectively. Similarly, the validation loss also exhibited notable convergence, stabilizing at 0.04 m and 0.045 m for NDPE without and with augmentation, respectively. This performance on the validation dataset suggests that the NDPE model has a high level of accuracy for unseen images. In particular, the model achieved an MSE loss of around 0.04 m in its single-frame predictions, demonstrating its feasibility and effectiveness in dock pose estimation. The results show that the model trained with the augmented dataset achieves a lower loss on the training set when the training step size is the same, suggesting that the NDPE learns more adequately how to predict dock positions. The results from the validation set surface that the model has strong robustness with a loss of around 0.04 on the unseen examples in training. The NDPE model was run on our onboard computer. It achieved a single-frame rollout frequency of 6 Hz.

\subsubsection{Effect of Data Efficiency}

\begin{figure}[htb]
    \centering
       \includegraphics[width=1\linewidth]{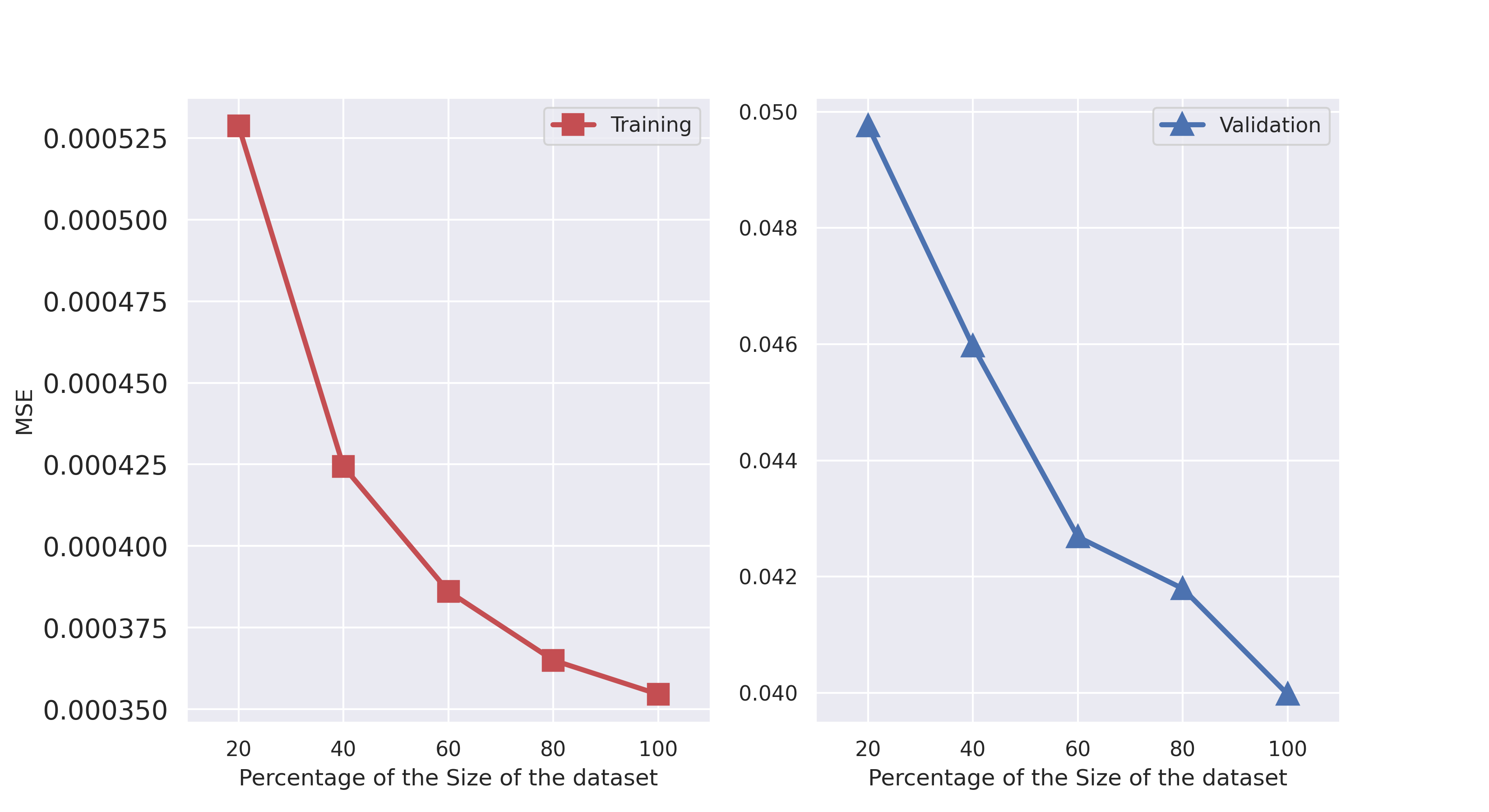}
    \label{fig:trans1}\hfill
	  \caption{The NDPE loss and data efficiency.}
	  \label{fig:data-efficiecny} 
\end{figure}

In Figure \ref{fig:data-efficiecny}, we evaluated the data efficiency of our proposed model for the autonomous docking task. We considered 400, 800, 1200, 1600 and 2000 dataset sizes for training and evaluation on a validation set. With a training data size of 400, the model training loss was 0.000525 and the validation set loss was around 0.00041. When we kept increasing the size of the training data, there was a clear trend of decreasing error in both the training and validation sets. At a data size of around 1200, the model training and validation set loss decreases begin to slow down. After that, the loss decreasing rate tends to be the same.

\newpage
\subsubsection{Effect of USV-Dock Distance for a Single-frame Prediction}

We were interested in the relationship between the model loss and the distance of the USV from the dock. Initially, we hypothesized that the closer the USV is to the dock, the smaller the model inference loss. 

\begin{figure}[htp]
    \centering
    \includegraphics[width=1\linewidth]{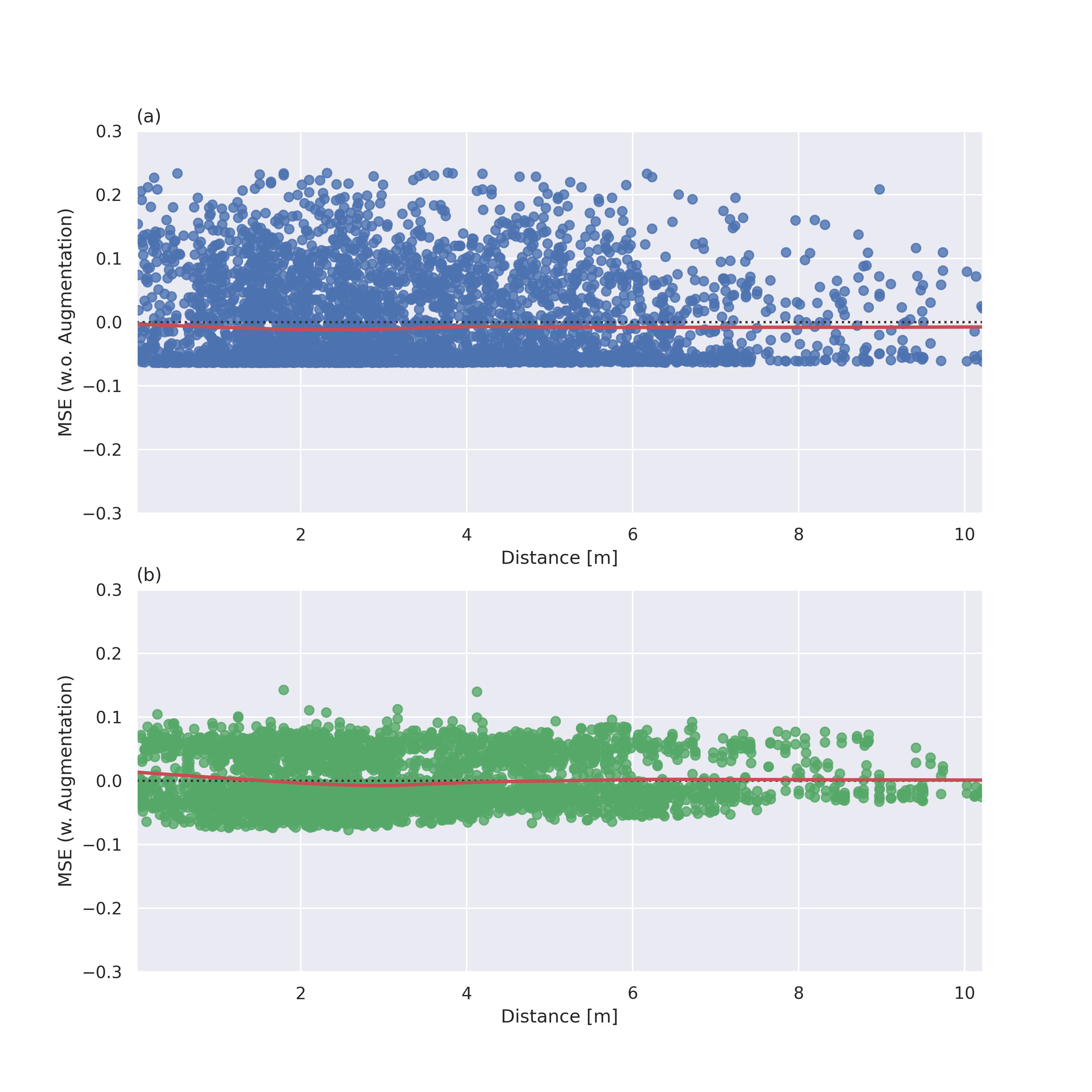}
    \caption{Distance-MSE relation for single-frame prediction.}
    \label{fig:distance}
\end{figure}

Figure \ref{fig:distance} reveals that its slope of the red regression line is almost zero, which indicates that the model's loss does not have a strong correlation with the distance of the USV from the dock. This also simultaneously shows that the accuracy of our model within 10 meters of the docking point for USVs is satisfactory the vast majority of the time. This implies that for most GPS-based navigation USVs a 10-m accuracy can be achieved, after which visual servo docking is performed, demonstrating the potential of our framework to integrate with existing systems. When we train with augmented data, the MSE of the NDPE model with distance is more centrally distributed around 0.1. When trained without using augmented data, the MSE of the NDPE model with distance change is more centrally distributed around 0.3. This shows that training with augmented data allows the model to have higher accuracy and robust to the distance perturbation.

\subsubsection{Effect of USV Velocity for Single-frame Prediction}

In the visual servo experiment, the speed of the USV may significantly influence the accuracy of visual docking. To investigate this, we analyzed the impact of USV velocity on the accuracy of single-frame predictions.

\begin{figure}[htp]
    \centering
    \includegraphics[width=1\linewidth]{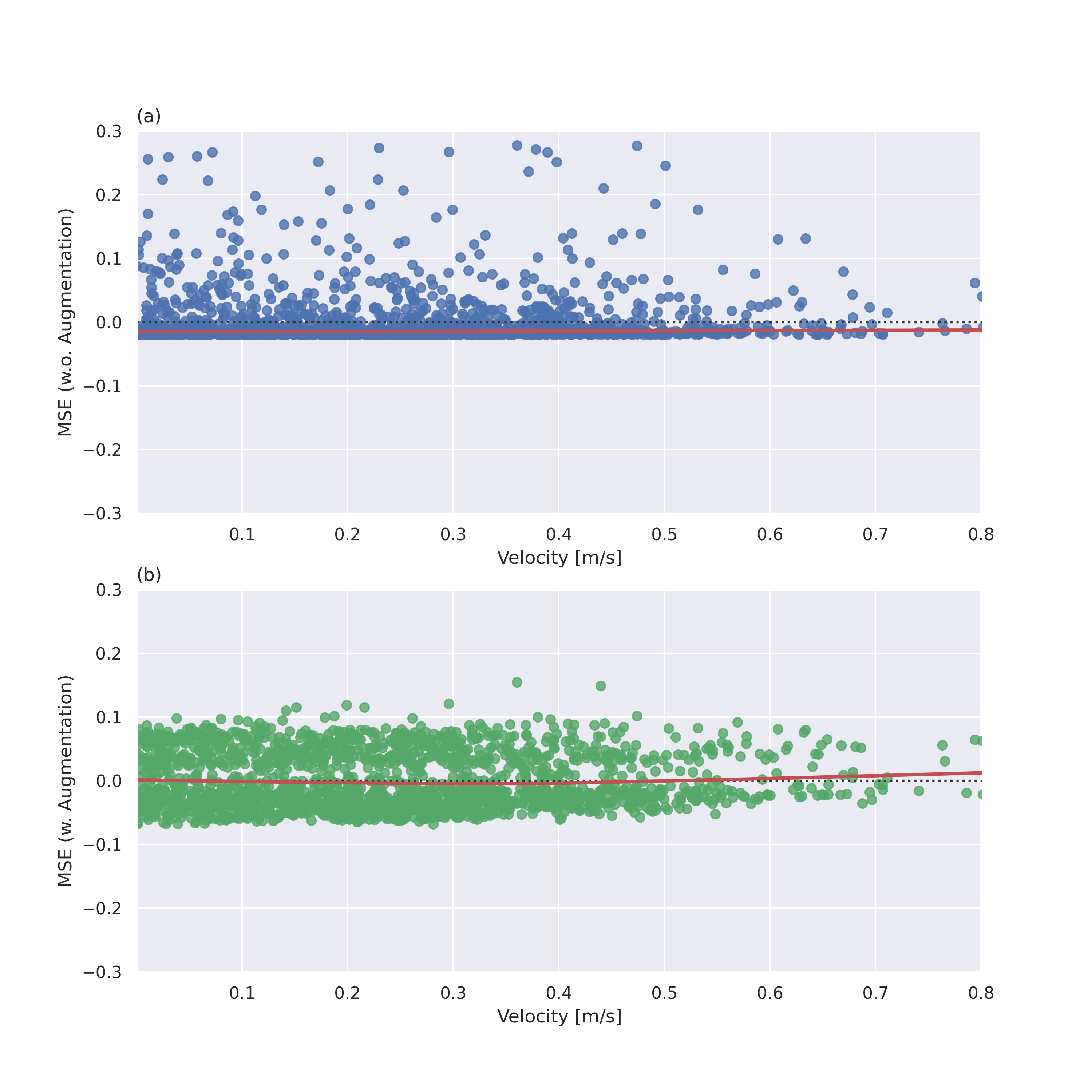}
    \caption{Velocity-MSE relation for single-frame prediction.}
    \label{fig:vel}
\end{figure}

Figure \ref{fig:vel} shows that the slope of the red regression line is nearly flat, indicating that the model's loss does not have a strong correlation with the velocity of the USV. This suggests that the accuracy of our model remains consistent across a range of velocities, ensuring reliable performance during visual docking. This finding highlights the potential of our framework to be integrated with visual servo docking is performed after reaching a certain velocity threshold between 0 to 0.8 m/s. When trained with or without augmented data, the MSE of the NDPE model is distributed around 0.1 m at various velocities. However, when trained with augmented data, the MSE is more uniformly distributed with respect to velocity variations. This result indicates that the augmented NDPE model is more responsive to changes in USV velocity, yet still maintains accuracy within the 0.1 m range.

\subsection{Experiments on Visual-Servo}
\label{subsec:vs-exp}

In order to validate the feasibility of the entire framework, we conducted 20 trials of real-world autonomous docking tasks across two different water scenarios. The experiment was conducted under average weather conditions characterized by a mean temperature of 26.3 °C and an average wind speed of 3.36 m/s. The same NDPE model was used for each experiment. The initial states were different in each trial. Then we evaluated the success rate of the proposed system.

\begin{figure*}[t]
    \centering
    \includegraphics[width=1\linewidth]{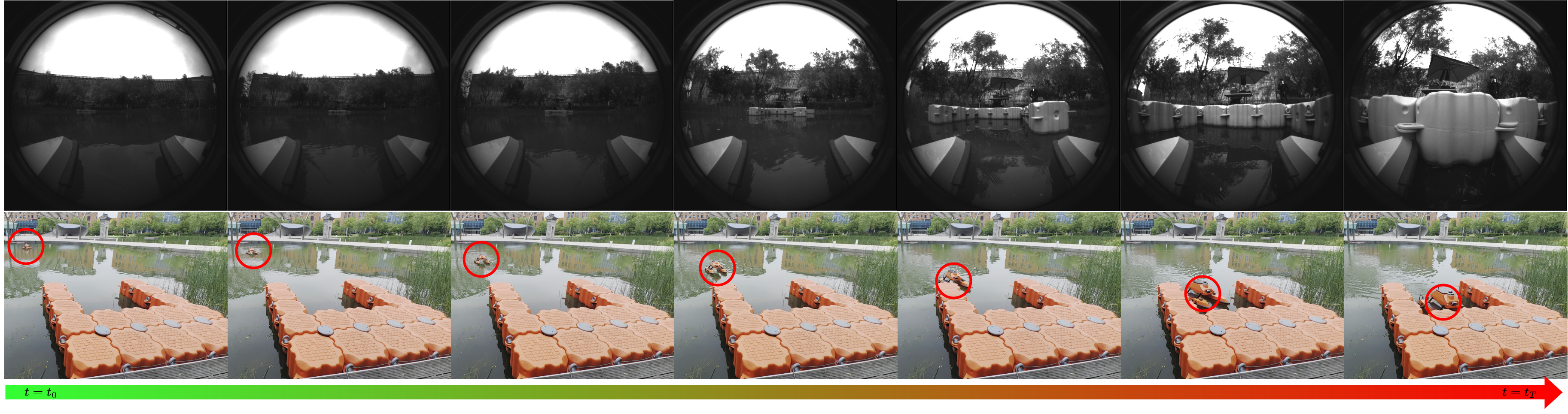}
    \caption{A sequence of first-person and third-person images of an trail.}
    \label{fig:real-exp}
\end{figure*}

Multiple experimental results indicated the feasibility of our framework for real USVs. In the 20 trials conducted, our proposed framework successfully achieved precise autonomous docking, effectively demonstrating its visual servoing capabilities in these real-world scenarios. Figure \ref{fig:real-exp} showed a sequence of images from the fisheye camera and the third-person view camera. In 18 of these cases, perfect docking was achieved and the USV entered the docking area, but there were also 2 cases where the framework failed to achieve perfect docking but successfully approached the dock. The reason for this result is that the initial angle of the USV is large and the angular velocity of the USV is too large at the beginning, resulting in the controller not being able to track the desired position and heading well during approaching. In addition, external factors in the real environment, such as wind and waves, and the offset of the dock position, also brought perturbations to each experiment, but our framework is closed-loop control, so accurate autonomous docking can still be achieved in most cases. In both experiments where there were failures, we noted that the NDPE results were still valid. Since the USV is a complex underactuated system, our controller did not perfectly execute some sharp turns. For example, the USV is quite near the dock and there was a large yaw error. The experiments showed that when the NDPE model is integrated into a visual servoing framework, visual docking can be effectively achieved in a wider range of scenarios and pre-docking poses. We found our implementations empirically achieved effective and precise results in real-world water environments.

\subsection{Discussions}

\subsubsection{Generalisability}

The ability of the system to generalize in the real world is critical for reliable deployment over long periods of time. Enhancing the system's ability to consistently perform autonomous docking in a variety of scenarios and environmental conditions enables truly unmanned, automated deployments. In computer vision, different water colours, lighting conditions and other maritime factors play an important role. We enhance generalisability capabilities through data augmentation. So that the system can be effective in different natural lighting, and in situations where motion brings image blur. In addition, to enhance the generalisability of our proposed module for different configurations of USVs, our pipeline also decouples the dock posture estimation from the motion controller. As a result, distinct configurations of USVs can use the same NDPE module for dock pose estimation. The lower-level controls can be adapted by the manufacturer or by a professional motion control engineer. In the future, in terms of data efficiency, data pairs can also be generated using realistic simulation techniques such as Unity or Omniverse.

\subsubsection{Explainability and Safety}

In contrast to black-box and end-to-end approaches, the learning-based component NDPE is only used for docking posture estimation and works in coordination with the hand-crafted low-level controller, thus the proposed pipeline design is interpretable and safe in most cases. This approach improves the safety of the pipeline data flow before executing control commands. The user can monitor the pose estimation values in real-time and abort the task anytime there is a fit of abnormal values.

\section{Conclusions and Future Works}\label{section conclusion}

This study successfully designed, developed and validated a self-supervised learning method for the visual docking of USVs. The NDPE achieved accurate dock pose estimation at the centimeter level across various dock positions. The proposed pipeline was rigorously tested in real-world environments, and the experimental results showed that our framework is viable for autonomous docking tasks with a high success rate. Our well-designed pipeline demonstrates the practical applicability in fully autonomous USV deployments.

Future work will focus on improving environmental adaptability through adaptive learning algorithms and online training techniques while enhancing control precision by exploring advanced strategies like Model Predictive Control (MPC) and Reinforcement Learning (RL). These improvements aim to ensure robust and reliable autonomous docking in challenging water environments.

\section*{Acknowledgments}
This research was funded by the Suzhou Science and Technology Project (SYG202122), Suzhou Municipal Key Laboratory for Intelligent Virtual Engineering (SZS2022004), the Research Development Fund of XJTLU (RDF-19-02-23) and Jiangsu Engineering Research Center for Data Science and Cognitive Computing.

\bibliographystyle{IEEEtran}
\bibliography{ref}
\end{document}